\documentclass{article} %
\usepackage{iclr2027_conference,times}

\usepackage{amsmath,amsfonts,bm}

\def\eqref#1{equation~\ref{#1}}

\def\1{\bm{1}}

\DeclareMathAlphabet{\mathsfit}{\encodingdefault}{\sfdefault}{m}{sl}
\SetMathAlphabet{\mathsfit}{bold}{\encodingdefault}{\sfdefault}{bx}{n}

\usepackage{wrapfig}
\usepackage{hyperref}
\usepackage{graphicx}
\usepackage{subcaption}
\usepackage{url}
\usepackage{graphicx}
\usepackage{booktabs}
\usepackage{multirow}
\usepackage{xcolor}
\usepackage{amsmath}
\usepackage{amssymb}
\usepackage{amsthm}
\usepackage{array}
\usepackage{tabularx}
\usepackage{tikz}  %

\definecolor{draftorange}{RGB}{213,94,0}

\usepackage{amsmath}
\usepackage{float}
\usepackage[breakable,skins]{tcolorbox}  %
\usepackage{tabularx}                    %
\usepackage{booktabs}                    %
\usepackage{array}                       %
\usepackage{xcolor}
\usepackage{soul}      %
\definecolor{hlred}{RGB}{250,214,214}
\definecolor{hlgreen}{RGB}{208,236,212}
\definecolor{hlgray}{RGB}{222,222,222}
\providecommand{\redt}[1]{{\sethlcolor{hlred}\hl{#1}}}
\providecommand{\greent}[1]{{\sethlcolor{hlgreen}\hl{#1}}}
\providecommand{\orig}[1]{{\sethlcolor{hlgray}\hl{#1}}}
\usepackage{booktabs}  %
\usepackage{array}
\usepackage{tabularx}  %
\usepackage{xspace}    %
\providecommand{\methodName}{\MitigateName}

\providecommand{\methodName}{\MitigateName}
\providecommand{\scislop}{SciSlop\xspace}

\definecolor{LightBlue}{RGB}{220,230,250}
\definecolor{LightRed}{RGB}{250,220,220}

\definecolor{Red}{rgb}{0.768, 0.054, 0.054}
\definecolor{Blue}{rgb}{0.152, 0.294, 0.925}
\definecolor{Green}{rgb}{0,0.4,0.7}

\usepackage{colortbl}
\definecolor{detcolor}{HTML}{2F4A8A}   %
\definecolor{revcolor}{HTML}{B5652B}   %
\definecolor{ourscolor}{HTML}{2C4FC4}  %
\definecolor{humancolor}{HTML}{4F7F5A} %
\definecolor{heatblue}{HTML}{4F8FA8}   %
\definecolor{heatred}{HTML}{C0505A}    %
\definecolor{linkred}{HTML}{A8323E}
\definecolor{citeteal}{HTML}{1F6F78}
\colorlet{oursbg}{ourscolor!7}
\colorlet{headbg}{black!6}
\newcommand{\oursrow}{\rowcolor{oursbg}}
\newcommand{\headrow}{\rowcolor{headbg}}
\newcommand{\famhead}[3]{\multicolumn{#1}{@{}l}{\color{#2}\textit{#3}}}

\hypersetup{
    colorlinks=true,
    citecolor=citeteal,
    linkcolor=linkred,
    urlcolor=ourscolor,
}
\usepackage{xurl} %

\usepackage{xspace}
\usepackage{enumitem}
\newcommand{\benchmarkName}{\textsc{SciSlopBench}\xspace}
\newcommand{\MitigateName}{\textsc{SciSlopHarness}\xspace}

\newcommand{\linedlabel}[2]{%
  {\color{#1}%
    \setbox0=\hbox{\texttt{#2}}%
    \rule[\dimexpr\ht0+1pt]{\wd0}{0.5pt}%
    \kern-\wd0%
    \texttt{#2}%
    \kern-\wd0%
    \rule[-2pt]{\wd0}{0.5pt}%
  }%
}

\definecolor{structurecolor}{HTML}{95627A}  %
\definecolor{argumentcolor}{HTML}{E4959E}   %
\definecolor{artifactcolor}{HTML}{6D8A96}   %
\colorlet{structurebg}{structurecolor!9}
\colorlet{argumentbg}{argumentcolor!13}
\colorlet{artifactbg}{artifactcolor!11}

\newcommand{\planetag}[2]{\tikz[baseline=(t.base)]{\node[inner xsep=3.4pt,inner ysep=1.6pt,
  rounded corners=3pt,draw=#1!85!black,line width=0.5pt,text=#1!85!black,
  font=\usefont{OT1}{cmss}{bx}{n}\fontsize{7.2}{7.8}\selectfont] (t) {\MakeUppercase{#2}};}}
\newcommand{\structure}{\planetag{structurecolor}{Structure}\xspace}
\newcommand{\argument}{\planetag{argumentcolor}{Argument}\xspace}
\newcommand{\artifacts}{\planetag{artifactcolor}{Artifacts}\xspace}

\title{\textit{Science or Slop?}: Benchmarking and Mitigating Scientific Slop in AI-Generated Papers}

\author{\begin{minipage}{\dimexpr\textwidth-2\tabcolsep\relax}\normalfont\centering
\vspace*{0.2in}
\textbf{Yerim Oh$^{1}$ \quad Young-Jun Lee$^{2}$ \quad Jaewoo Ahn$^{1}$ \quad Gunhee Kim$^{1}$ \quad Dongyeop Kang$^{2}$} \\
$^{1}$Seoul National University \quad $^{2}$University of Minnesota \\
\texttt{yerim.oh@vision.snu.ac.kr} \quad \texttt{gunhee@snu.ac.kr} \quad
\texttt{dongyeop@umn.edu} \\[3pt]
\raisebox{-0.2em}{\includegraphics[height=1.05em]{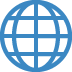}}\ \textbf{Website:} \url{https://yerimoh.github.io/scientific-slop-demo/}
\end{minipage}
}

\begin{document}

\maketitle
\vspace{-0.1in}

\begin{abstract}
AI-generated content, often called AI slop, is increasingly common everywhere, particularly in academia.
Slop in AI-generated scientific papers, however, has more complex patterns that cannot be easily detected by existing token-based AI detectors.
Each part of such a paper looks plausible while the scientific reasoning that connects the parts breaks down, which can mislead how readers assess the work.
We benchmark these failures as \textbf{scientific slop} through six measures across \emph{Structure}, \emph{Argument}, and \emph{Artifacts}. We construct \textbf{\benchmarkName} with 390 AI-generated papers, mostly in computer science but spanning the life, social, and natural sciences, each paired with a human-written paper matched by research problem and contribution type. Our measures identify the AI paper in each pair with 85.9\% accuracy, compared with 68.7\% for Binoculars. Higher scientific slop accompanies lower ICLR ratings and distinguishes rejected from accepted papers above chance in every year from 2017 to 2025. Reducing these patterns, however, is not as simple as directly optimizing the measures. We therefore propose \textbf{\MitigateName}, a harness-level framework that guides a fixed LLM to revise slop only where the experiment records support the change. 
While standard revisions leave residual slop and direct slop-aware prompting triggers reward hacking, \MitigateName reduces the remaining AI--human gap by 63\% over the strongest revision baseline without requiring human reference targets. Overall, we demonstrate that AI-generated scientific papers leave fundamental traces in their global reasoning, and that responsible mitigation demands strict evidentiary grounding rather than mere prose refinement.

\end{abstract}

\section{Introduction}
\label{sec:introduction}

As LLM-generated content fills news feeds, books, and social media, the term \emph{AI slop} has come to name this low-cost, high-volume output~\citep{thorp2026resisting}. AI output that looks finished but lacks substance shifts the verification burden to the recipient, forcing them to review and revise the content, and raises concerns of misuse in journalism, education, workplaces, and academia~\citep{wang2024m4}. 
AI slop is increasingly common in academia, in both papers and peer reviews~\citep{liang2025quantifying,kobak2025delving,liang2024monitoring}. 
Paper submission platforms such as arXiv have revised their policies to restrict survey papers~\citep{arxiv2025reviewpolicy}, and conference organizers have begun screening submissions using AI detectors. Yet, a scientific paper is more than its \emph{prose}. 

These concerns have made detecting AI-generated content increasingly important. 
AI agents are now used to draft and refine scientific papers, from researcher-guided workflows~\citep{schmidgall2025agent} to systems such as AI Scientist~\citep{lu2024aiscientist,yamada2025ai} and FARS~\citep{tang2026fars}. Such assistance can reduce writing time while also increasing the need for verification, as generated scientific text can appear credible even when the underlying data are fabricated~\citep{kacena2024use, gao2023comparing}. Existing detection methods address these concerns by identifying AI-generated content through learned textual features or token probabilities~\citep{emi2024pangram,mitchell2023detectgpt,hans2024binoculars,ma2026zero}.
For scientific papers, however, token-level probabilities alone are insufficient. 

\begin{figure}[t]
\centering
\includegraphics[width=\textwidth]{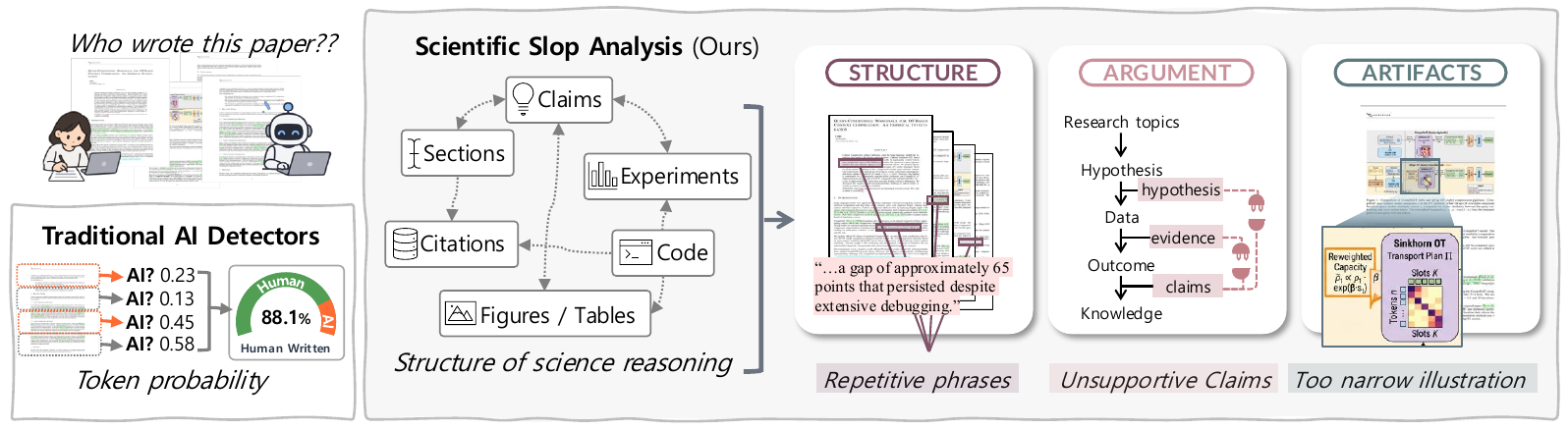}

\caption{Scientific slop captures failures beyond token-level AI detection.
While traditional detectors assess local text, scientific slop analysis examines how scientific reasoning connects, revealing paper-level failures that plausible prose can conceal.}
\label{fig:teaser}
\end{figure}

\textbf{Challenges.}
We define \textbf{scientific slop} as recurring breakdowns in how scientific reasoning connects across a paper (Figure~\ref{fig:teaser}).
Each part of a paper can look plausible in isolation, so such breakdowns are invisible to token-level detectors and can only be identified or repaired at the level of the whole paper.
To benchmark and mitigate scientific slop, this paper addresses three challenges.
\textbf{First, token-level metrics fail to capture how scientific reasoning connects across a paper.} Sections, claims, citations, evidence, and artifacts can each appear plausible while the relationships among them break down. We repeatedly observe such failures in end-to-end AI-generated papers, and ICLR reviewers already penalize them even in human-written submissions. %
\textbf{Second, detection of these patterns remains unmeasured.} Existing test sets label only the text, so the extent to which detectors, including LLMs that read the entire paper, identify these patterns has never been measured. 
\textbf{Third, these patterns are difficult to mitigate reliably.} Whereas token-level signals can be removed by paraphrasing~\citep{krishna2023paraphrasing}, repairing these patterns requires restoring the missing relations without changing the underlying science.

To systematically study scientific slop in AI-generated papers, we construct \textbf{\benchmarkName}, the first benchmark to evaluate detection in full scientific papers rather than standalone prose.
It pairs 390 AI-generated papers from FARS and Agents4Science 2025~\citep{agents4science2025} with human-written papers on the same research problem and of the same contribution type, mostly in computer science but also spanning the life, social, and natural sciences.
Every human-written counterpart was accepted at a top-tier venue, so a system that distinguishes the two must look beyond topic, genre, and prose quality.
To mitigate the identified scientific slop, we introduce \textbf{\MitigateName}, which translates slop diagnoses into iterative revision guidance for an LLM. In each round, it checks edits against the original manuscript and experiment records, then uses review feedback to repair remaining gaps in scientific reasoning.

\textbf{Contributions.}
Our results formalize scientific slop as paper-level breakdowns in scientific reasoning and develop measures to quantify it. These measures reduce detection error by 55\% compared with Binoculars~\citep{hans2024binoculars}, the strongest token-level baseline.
Moreover, measures of scientific slop align with human evaluations: AI-likeness drops from 26\% to 12\% as review scores rise from 2 to 9. These measures also distinguish rejected from accepted papers in every ICLR year from 2017 to 2025.
For mitigation, general revision leaves slop unresolved, while direct slop-aware revision lowers the scores without fixing the reasoning. \methodName, which repairs scientific reasoning against grounding evidence, instead moves all six measures toward human levels without using human targets. 
These results show that AI-generated papers leave measurable traces in how scientific reasoning connects across a paper, and that grounding revision in scientific records enables these traces to guide scientific repair.
Our contributions are:
\begin{itemize}[leftmargin=2em, itemsep=1pt, parsep=0pt, topsep=1pt]
    \item We define and measure scientific slop through six types of breakdown in scientific reasoning across \structure, \argument, and \artifacts. Individual measures achieve a pairwise accuracy of up to 0.905 in distinguishing AI-generated from human-written papers.
    \item We construct \benchmarkName to evaluate AI detection based on scientific reasoning across a whole paper rather than standalone prose. The strongest token-level detector, Binoculars, achieves a pairwise accuracy of only 0.687, while full-text AI reviewers achieve at most 0.685.  %
    \item We introduce \MitigateName{}, which verifies repairs against scientific records rather than slop scores. It reduces the remaining AI–human gap by 63\% over the strongest revision baseline, while general revision leaves scientific slop and direct slop-aware revision overcorrects.
\end{itemize}

\section{Related Work}
\label{sec:related}

\textbf{Detecting machine generated text.}
Detection methods read local text spans. DetectGPT measures probability curvature \citep{mitchell2023detectgpt}, Binoculars compares model perplexity \citep{hans2024binoculars}, NTS measures sensitivity to sampling temperature \citep{ma2026zero}, and Pangram trains classifiers on generated text \citep{emi2024pangram}. Watermarking and paraphrasing studies also work on surface signals \citep{kirchenbauer2023watermark,krishna2023paraphrasing}. Unlike these tools, our measures read relations across a whole paper rather than local text spans.

\textbf{Automated review and revision.}
Recent systems generate critiques or revise drafts directly. The AI Scientist evaluates its own papers with a simulated review process \citep{lu2024aiscientist}, CycleReviewer learns from real peer reviews \citep{weng2025cycleresearcher}, and OpenReviewer and DeepReview treat review generation as a primary task \citep{idahl2025openreviewer,zhu2025deepreview}. Self-Refine closes the loop by having one model critique and revise its own output \citep{madaan2023selfrefine}. RefineBench scores refinement against checklists and finds that models improve little without feedback \citep{lee2026refinebench}. REPAIR grounds each refinement round in documents verified through scientific APIs \citep{oh2026repair}.
However, these systems offer general feedback without isolating specific recurring patterns. We use the AI Scientist reviewer and CycleReviewer as detection baselines and a reviewer-based refinement loop as a revision baseline, and ask whether general review and revision removes the flaws we define.

\textbf{Measurement under active removal.}
Recent work improves an LLM system by changing the information, checks, and tools around a fixed model rather than the model itself~\citep{lee2026meta}. 
Meta$^n$ applies one fixed improvement operation recursively to its own outputs~\citep{kim2026metan}, and Evolution Fine-Tuning trains the model on evolutionary search trajectories~\citep{lee2026evolution}.
\methodName{} follows this design, keeping the editor fixed and adding located findings before the edit and a record-grounded review after it. Mutation testing~\citep{demillo1978hints,jia2011mutation} and adversarial stylometry~\citep{brennan2012stylometry,potthast2016obfuscation} judge a property by trying to remove it, and our slop-aware baseline plays that role.

\section{Defining and Measuring scientific slop}

\label{sec:slop}
\begin{table}[t]
\centering
\caption{Measurement specification of \scislop patterns. Each score is the share of units that show the pattern, written as numerator\,/\,denominator (Eq.~\ref{eq:slop-item-score}). The last column states what a score of 1 means. Full rules are in Appendix~\ref{app:measurers}.}
\label{tab:slop-registry}
\label{tab:mold-registry}
\vspace{-5pt}
\scriptsize
\setlength{\tabcolsep}{2.0pt}
\renewcommand{\arraystretch}{1.}
\setlength{\aboverulesep}{0pt}\setlength{\belowrulesep}{0pt}
\newcommand{\schpage}[1]{\ifnum\pdfstrcmp{#1}{xsec_ref}=0 1\else\ifnum\pdfstrcmp{#1}{macro_redund}=0 2\else\ifnum\pdfstrcmp{#1}{argument_graph}=0 3\else\ifnum\pdfstrcmp{#1}{citation}=0 4\else\ifnum\pdfstrcmp{#1}{fig_exposition}=0 5\else 6\fi\fi\fi\fi\fi}
\newcommand{\schfig}[1]{\raisebox{0pt}[\dimexpr\height+0.5mm\relax][0.5mm]{\includegraphics[height=0.9cm,page=\schpage{#1}]{figures/main_figures/fig_tab1_pictograms.pdf}}}
\providecommand{\bandsep}{\arrayrulecolor{white}\specialrule{0.8pt}{0pt}{0pt}\arrayrulecolor{black}}
\newcommand{\bandtag}[2]{\raisebox{0.5cm}[0pt][0pt]{\setlength{\fboxsep}{1.5pt}\colorbox{#1}{\scalebox{.65}{#2}}}}  %
\newcommand{\scorefrac}[3][black!45]{{\color{#1}\ensuremath{\dfrac{\parbox{\dimexpr\linewidth-2.5pt}{\centering\color{black}#2}}{\parbox{\dimexpr\linewidth-2.5pt}{\centering\color{black}#3}}}}}  %
\renewcommand{\tabularxcolumn}[1]{>{\centering\arraybackslash}m{#1}}
\begin{tabularx}{\linewidth}{@{}>{\centering\arraybackslash}m{1.3cm} >{\raggedright\arraybackslash}m{1.4cm} >{\centering\arraybackslash}m{2.2cm} >{\raggedright\arraybackslash}m{2.32cm} X >{\raggedright\arraybackslash}m{2.72cm}@{}}
\toprule
\headrow & Item & Illustration & Unit of analysis $U_j(p)$ & Score $s_j(p)$ & What $s_j=1$ means \\
\midrule
\rowcolor{structurebg} & Cross-section references & \schfig{xsec_ref} & Body sections and labeled objects & \scorefrac[structurecolor!55]{objects never referenced outside their own section}{all objects} & {No object is ever referred to from another section.} \\
\bandsep
\rowcolor{structurebg} \bandtag{structurebg}{\structure{}} & Macro redundancy & \schfig{macro_redund} & Sentences with at least eight tokens & \scorefrac[structurecolor!55]{sentences half or more copied as 8-grams from an earlier section}{all sentences} & {Every sentence mostly repeats earlier sections.} \\
\midrule
\rowcolor{argumentbg} & Argument graph & \schfig{argument_graph} & Key claims in the Introduction & \scorefrac[argumentcolor!70]{shallow claims: nothing earlier leads up to them}{all key claims} & {The argument is flat. Every claim stands alone with no build-up behind it.} \\
\bandsep
\rowcolor{argumentbg} \bandtag{argumentbg}{\argument{}} & Citation isolation & \schfig{citation} & Citation sentences (Intro., Related Work) & \scorefrac[argumentcolor!70]{citations not grouped, compared, or related to other works}{all citations} & {Every citation stands alone.} \\
\midrule
\rowcolor{artifactbg} & Figure exposition & \schfig{fig_exposition} & Content types in method figures & \scorefrac[artifactcolor!55]{content types not needed to show the method}{all content types in the figure} & {Nothing in the figure shows the method itself.} \\
\bandsep
\rowcolor{artifactbg} \bandtag{artifactbg}{\artifacts{}} & Evidence gap & \schfig{evidence_gap} & Papers with a body result table & \scorefrac[artifactcolor!55]{papers showing no concrete input, output, or case}{all papers} & {The paper gives no concrete example at all.} \\
\bottomrule
\end{tabularx}
\end{table}

\subsection{Scientific Slop: Definition and Measures}

\citet{10.1145/3786777} identify superficial competence as a characteristic of AI slop, whose apparent quality conceals a lack of substance.
In scientific papers, we examine this gap in the connections needed to organize the study, substantiate its claims, and make its methods and evidence inspectable.
We define \textbf{scientific slop} as recurring failures in these connections that hinder readers from following the paper's scientific reasoning.
We group these failures into \structure, \argument, and \artifacts.
Table~\ref{tab:slop-registry} specifies six observable patterns and their counting rules.

\structure concerns how sections contribute to the paper’s scientific argument. \emph{Cross-section references} targets missing connections between them. Rhetorical structure theory explains how parts of a text are connected, while endophoric references make these connections visible to readers~\citep{mann1988rst,hyland2005metadiscourse}. We therefore measure sections and labeled objects that no other section refers to.
\emph{Macro redundancy} captures when later sections repeat earlier material instead of developing the scientific argument. Methods, results, and conclusions serve different argumentative roles~\citep{teufel2009towards}, so each section should add to rather than restate the argument. We measure such reuse with cross-section $n$-gram overlap~\citep{welleck2020unlikelihood}.

\argument evaluates how a paper justifies its claims and establishes its position among prior studies.
\emph{Argument graph} targets breaks in the reasoning that connects a paper’s claims to their support. Argument mining represents this reasoning through relations between claims and premises~\citep{stab2017parsing}. We measure how often a claim appears before its supporting context.
\emph{Citation isolation} captures prior work that is cited without explaining its role in the paper’s argument. Citations can show what a paper builds on, differs from, or compares against~\citep{teufel2006citation}. We measure citations that provide none of these relations.

\begin{figure}[t]
\centering
\includegraphics[width=\textwidth]{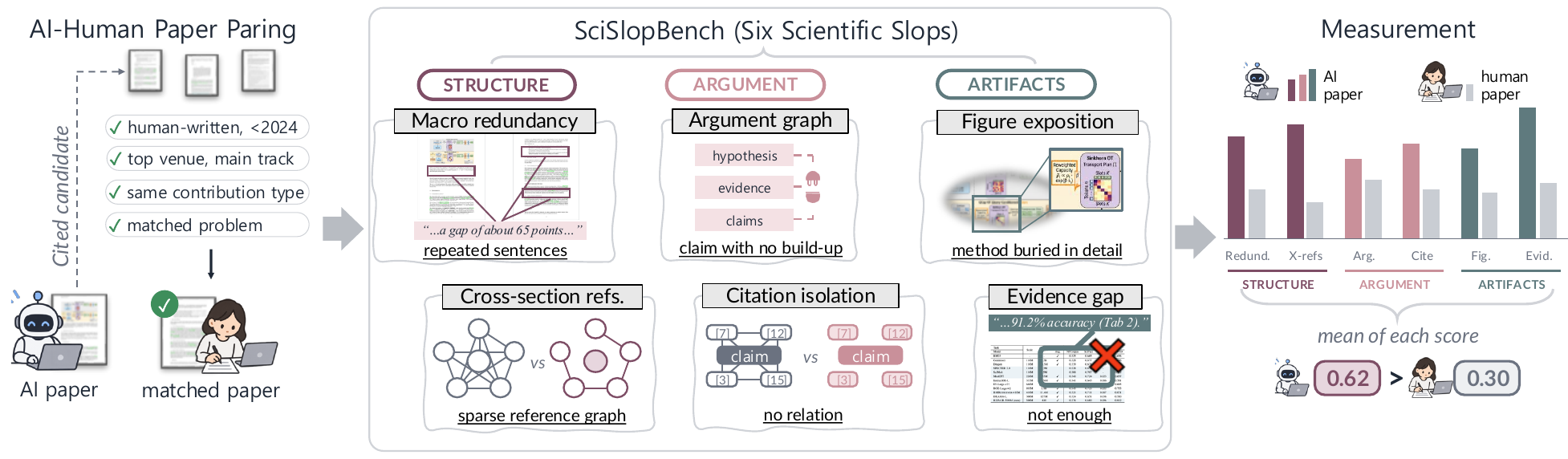}
\caption{\textbf{\benchmarkName{} pipeline.} We pair each AI-generated paper with a human-written paper matched by research problem and contribution type. We measure six scientific slop patterns in both papers, then compare the mean slop scores of each matched pair.}
\label{fig:pipeline}
\end{figure}

\artifacts examines whether readers can inspect how a method works and what its reported results represent.
\emph{Figure exposition} captures method diagrams that obscure how the method works.
Diagrams support inference by showing related information together~\citep{larkin1987diagram}, but unnecessary explanations can interfere with this role~\citep{mayer2003nine}.
We flag experimental details or claims about results that make the method's steps and connections harder to follow.
\emph{Evidence gap} occurs when a paper reports aggregate results but provides no concrete examples.
Without these examples, readers cannot inspect individual successes and failures to better understand model behavior~\citep{lipton2018troubling}.
We check for this gap in papers with result tables in the main text, looking for inputs, outputs, or cases throughout the paper and appendix.

For paper $p$ and item $j$, let $U_j(p)$ be the units of analysis and $h_j(u)$ indicate whether unit $u$ meets the counting criterion in Table~\ref{tab:slop-registry}. The item score is
\begin{equation}
    s_j(p)=\frac{1}{|U_j(p)|}\sum_{u\in U_j(p)}h_j(u).
    \label{eq:slop-item-score}
\end{equation}

\subsection{\benchmarkName}
\label{sec:SciSlopBench}

Existing AI text detection benchmarks~\citep{wang2024m4,li2024mage,dugan2024raid,wu2024detectrl} typically evaluate standalone prose, leaving relations among a paper's sections, tables, and code untested. To study scientific slop systematically, we introduce \benchmarkName{}, a benchmark for evaluating how well systems distinguish AI-generated papers from matched human-authored papers using whole-paper evidence. As shown in
Fig.~\ref{fig:pipeline}, \benchmarkName{} pairs AI-generated papers with human-written papers
matched by research problem and contribution type, gives each paper one slop score from the six patterns, and checks whether the human-written paper scores lower than its AI-generated counterpart.

\subsubsection{Benchmark Construction}
\label{sec:bench_construction}

Each instance in \benchmarkName{} pairs an AI-generated paper with a human-authored paper matched by paper type and topic. %
We evaluate whether these scores rank the human-authored paper above its AI-generated counterpart. The reference labels reflect how the papers were produced.

\paragraph{Dataset construction.} 
We draw AI-generated papers from FARS and Agents4Science 2025 submissions. FARS takes a research idea and produces a finished paper, running the experiments and writing the manuscript itself. For each FARS paper, we identify cited papers of the same paper type that meet our human-authorship screening criteria and have main-track acceptance at a top-tier venue. We then select the closest topical match, yielding 143 pairs matched on both the kind of contribution and the research topic. For Agents4Science, we also include papers available as PDFs, which we convert into a form suitable for slop measurement. Data collection, matching, and document-processing details are provided in Appendix~\ref{app:data_detail}.

\paragraph{Benchmark statistics.}

\benchmarkName combines 143 matched pairs from FARS with 247 pairs from Agents4Science 2025 submissions, spanning computer and data sciences, life and health sciences, social sciences, natural sciences, and other fields. We report slop measures as normalized rates to account for differences in paper length and the number of measured units. Source-specific composition, acceptance counts, and detailed topic distributions are provided in Appendix~\ref{app:data_detail}.

\paragraph{Evaluation metrics.}
We use PairAcc and AUROC as our main evaluation metrics. PairAcc measures how often a system ranks the human-authored paper above its AI-generated counterpart, with score directions fixed in advance and ties receiving half credit. AUROC measures how well the scores distinguish the two classes across classification thresholds. We also report the true-positive rate at a 5\% false-positive rate (TPR), the share of AI papers flagged when the threshold lets through 5\% of human papers. For the ICLR analysis in Section~\ref{sec:slop_provenance_quality}, each system's score is turned into an AI probability by a logistic fit on the FARS pairs (Appendix~\ref{app:measurers_review}).

\begin{figure}[t]
\centering
\includegraphics[width=\textwidth]{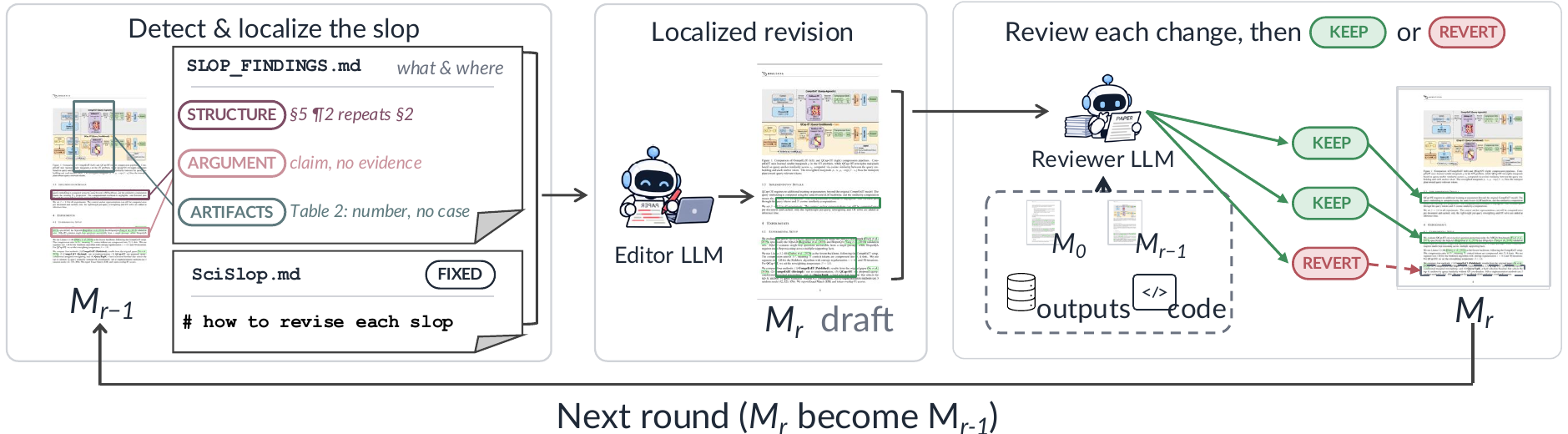}
\caption{\textbf{\MitigateName{} pipeline.} We locate scientific slop instances in the manuscript and use them to guide localized revision. We then review each proposed change against the original manuscript, experiment outputs, and code, retain only supported changes, and repeat the process on the remaining slop.}
\label{fig:scislop-harness}
\end{figure}

\section{\MitigateName: Mitigating Scientific Slop}
\label{sec:mitigation}

This section describes \MitigateName{}, our method for reducing scientific slop by repairing the
connections in scientific reasoning. As shown in Fig.~\ref{fig:scislop-harness}, \MitigateName{} locates
scientific slop, uses the detected instances to guide manuscript revision, and reviews each proposed
change against the original manuscript and the paper's experiment records before retaining it. The
remaining slop and review feedback then guide the next revision round.

\paragraph{Task.}
Given an original manuscript $M_0$ and its experiment records, the result files its code produced, \MitigateName{} revises the manuscript for up to three rounds.
Following harness-level optimization~\citep{lee2026meta}, we keep the revision model fixed and change only the information and checks around it.
At round $r$, the revision model receives $M_{r-1}$ with \texttt{SciSlop.md}, which describes how each type of scientific slop should be revised.
It also receives \texttt{SLOP\_FINDINGS.md}, which lists the detected problems and their locations.
The revision model edits the manuscript to produce a draft $\widetilde{M}_r$.
A review model then compares each change with $M_{r-1}$, $M_0$, experiment outputs, and the code.
Changes consistent with the reported results, existing evidence, and cited work are kept in $M_r$.
The remaining changes are reverted to their version in $M_{r-1}$, so each round proceeds as $M_{r-1} \overset{\mathrm{revision}}{\longrightarrow} \widetilde{M}_r \overset{\mathrm{review}}{\longrightarrow} M_r$.
Appendix~\ref{app:scislop-harness} provides implementation details.

\subsection{Localized Revision}
\label{sec:Localized}

We first locate where a connection in scientific reasoning breaks down so that the revision model knows what part of the manuscript to revise.
For each detected problem, the model receives its location and guidance on how to revise it.
The locations come from measurements of the manuscript $M_{r-1}$ and are stored in \texttt{SLOP\_FINDINGS.md}.
The revision guidance is fixed across manuscripts and stored in \texttt{SciSlop.md}.
The model uses both to edit the \LaTeX{} source and produce $\widetilde{M}_r$.

We use these findings to guide revision, not to require every scientific slop measure to decrease.
The model can leave a finding unchanged when the suggested connection is not needed in the paper.
For example, it adds a cross-section reference only when another section actually uses the referenced object.
We therefore provide the findings and revision guidance without a target score.

\subsection{Reviewing Proposed Revisions}
\label{sec:Reviewing}

Localized guidance tells the revision model what to change, but it cannot verify whether the new content is supported by the research.
We therefore review each proposed change before keeping it in the manuscript.
For each changed paragraph in $\widetilde{M}_r$, the review model checks the original manuscript, experiment outputs, and code for the claims, numbers, citations, and examples affected by the revision.
It keeps a change only when these sources support the revised content.
Otherwise, the paragraph is restored to its version in $M_{r-1}$.
The remaining changes form $M_r$.

The review checks that revisions preserve the paper's reported results, citations, and available evidence.
A changed result is reverted if it conflicts with the experiment outputs.
A new example is kept only when the same example exists in the available records.
Changes that remove citations or add unnecessary references are also reverted.

We repeat revision and review for up to three rounds.
We stop when no problems remain for revision or when review produces no further changes.
Before returning the final manuscript, we check that reported numbers and cited works are preserved and that citations and references remain valid.

\begin{table*}[t]
\centering
\begin{minipage}[t]{0.415\textwidth}
\vspace{0pt}
\centering
\footnotesize
\setlength{\tabcolsep}{3pt}
\renewcommand{\arraystretch}{1.2}
\setlength{\aboverulesep}{0pt}\setlength{\belowrulesep}{0pt} %
\caption{
Paper discrimination on \benchmarkName{}. TPR is at 5\% FPR, and bold and
underline mark the best two.
}
\begin{tabular}{@{}l rrr@{}}
\toprule
\headrow Method & PairAcc & AUROC & TPR \\
\midrule
\famhead{4}{black}{AI text detectors, prose view} \\
\hspace{0.6em}Binoculars     & \underline{0.687} & 0.683 & \underline{0.238} \\
\hspace{0.6em}DetectGPT      & 0.638 & 0.623 & 0.146 \\
\hspace{0.6em}NTS            & 0.626 & 0.607 & 0.123 \\
\addlinespace[3pt]
\famhead{4}{black}{Automated reviewers, full text} \\
\hspace{0.6em}CycleReviewer  & 0.685 & \underline{0.689} & 0.054 \\
\hspace{0.6em}AI Scientist   & 0.615 & 0.613 & 0.146 \\

\midrule
\oursrow \textbf{\scislop (ours)} & \textbf{0.859} & \textbf{0.854} & \textbf{0.272} \\
\bottomrule
\end{tabular}

\label{tab:main_results}
\end{minipage}

\hfill
\begin{minipage}[t]{0.563\textwidth}
\vspace{0pt}
\centering
\footnotesize
\setlength{\tabcolsep}{3pt}
\caption{
Scientific slop measures by plane. TPR is at 5\% FPR, a dagger marks a measure that needs a model, and the last row is the score of Table~\ref{tab:main_results}.
}
\renewcommand{\arraystretch}{1.19}
\setlength{\aboverulesep}{0pt}\setlength{\belowrulesep}{0pt} %
\begin{tabular}{@{}l l rrr@{}}
\toprule
\headrow Plane & Measure & PairAcc & AUROC & TPR \\
\midrule
\rowcolor{structurebg} \structure & Macro redundancy & 0.723 & 0.727 & \underline{0.467} \\
           \rowcolor{structurebg} & Cross-section refs. & \textbf{0.905} & \textbf{0.895} & \textbf{0.648} \\
\addlinespace[7.8pt]
\rowcolor{argumentbg} \argument & Argument graph$^\dagger$ & 0.586 & 0.584 & 0.159 \\
           \rowcolor{argumentbg} & Citation isolation & 0.793 & 0.782 & 0.444 \\
\addlinespace[7.8pt]
\rowcolor{artifactbg} \artifacts & Figure exposition$^\dagger$ & \underline{0.809} & \underline{0.786} & 0.364 \\
           \rowcolor{artifactbg} & Evidence gap & 0.764 & 0.765 & 0.000 \\
\midrule
\oursrow \multicolumn{2}{@{}l}{\textbf{\scislop (ours)}, mean of planes} & 0.859 & 0.854 & 0.272 \\
\bottomrule
\end{tabular}

\label{tab:slop_measures}
\end{minipage}

\end{table*}

\section{Experiments}
\label{sec:exp_dis}

\subsection{Experimental Setup}

\paragraph{Baselines.}
All existing methods assign a single score to each paper. We evaluate
Binoculars~\citep{hans2024binoculars}, DetectGPT~\citep{mitchell2023detectgpt} and NTS~\citep{ma2026zero} as AI text
detectors, and use the overall ratings from the AI Scientist
reviewer~\citep{lu2024aiscientist} and CycleReviewer~\citep{weng2025cycleresearcher} as
automated review baselines.
For revision, we compare four baselines that share the editor model and round protocol.
Base prompting, Claude Code, and reviewer-based refinement receive only a generic
improvement instruction, and we refer to these three as general revision.
Slop-aware revision instead receives explicit slop definitions
and detected locations, using the same editor and tool configuration
as reviewer-based refinement.
We compare these baselines with our \MitigateName, using the negative of the aggregate item score so
that higher values favor human authorship, as for the other methods.
Details of the revision baselines are provided in Appendix~\ref{app:baseline_detail},
and the distances behind the reported gap reduction in Appendix~\ref{app:scislop-gap}.

\paragraph{Implementation details.}
Text detectors read the prose view. Other methods read the paper source, with code access
provided where needed. Binoculars uses the Falcon-7B~\citep{almazrouei2023falcon} base and instruct
models, while DetectGPT uses t5-3b~\citep{raffel2020t5} to generate 100 perturbations. For inputs exceeding Falcon's
2048-token context limit, we score windows without overlap and average their scores.
The AI Scientist reviewer uses Qwen2.5-32B-Instruct~\citep{yang2024qwen25}, and CycleReviewer uses its released 8B
checkpoint. 

\subsection{Does Scientific Slop Mark Provenance, Quality, or Both?}
\label{sec:slop_provenance_quality}

We test whether scientific slop captures overlooked weaknesses already reflected in lower human review scores. We first assess how well slop measures identify AI-generated papers, then examine whether the same patterns matter to human peer review.

\paragraph{Scientific slop identifies AI-generated papers by what detectors and reviewers never read.}
Tables~\ref{tab:main_results} and~\ref{tab:slop_measures} show that
scientific slop identifies AI-generated papers more accurately than the listed
detectors and automated reviewers that read the full paper.
The best of each reaches 0.687 and 0.685, while scientific slop reaches 0.859, cutting the error by more than half. \emph{Cross-section references} alone reaches 0.905 without a model and, at 5\% FPR, flags 65\% of AI-generated papers against 24\% for Binoculars.
These results show that measuring patterns across a paper provides
an effective signal for AI detection on this corpus.
By defining these patterns as specific defects, scientific slop further
connects detection to concrete targets for revision.

\begin{figure}[t]
\centering
\captionsetup[subfigure]{font=footnotesize,skip=3pt,position=top}
\begin{minipage}{0.98\linewidth}
\centering
\begin{subfigure}[t]{0.2730\linewidth}
\centering
\caption{Review scores}
\label{fig:review-scores-rating}
\includegraphics[width=\linewidth]{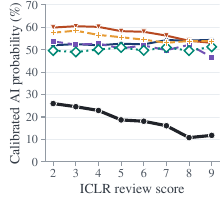}
\end{subfigure}\hfill
\begin{subfigure}[t]{0.4440\linewidth}
\centering
\caption{Review dimensions}
\label{fig:review-scores-dimensions}
\includegraphics[width=\linewidth]{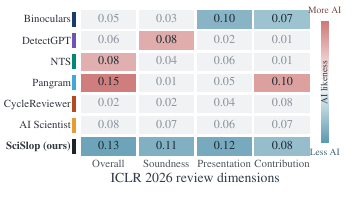}
\end{subfigure}\hfill
\begin{subfigure}[t]{0.2730\linewidth}
\centering
\caption{Acceptance decisions}
\label{fig:review-scores-acceptance}
\includegraphics[width=\linewidth]{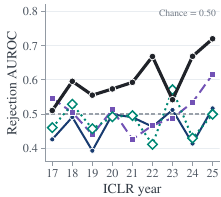}
\end{subfigure}
\par\vspace{2pt}
\includegraphics[width=\linewidth]{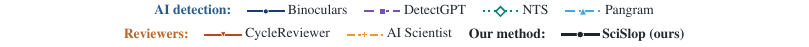}
\end{minipage}
\caption{AI likeness against ICLR review scores (a), assessment dimensions (b),
and annual acceptance decisions (c).
Panel (a) adjusts probabilities for differences between years, panel (b) shows
$|\rho|$ with \textcolor{heatblue}{blue} and \textcolor{heatred}{red} for significant negative and positive correlations
($p<0.05$) and gray for estimates that are not significant, and panel (c) uses
rejection AUROC with $0.5$ indicating chance.}
\label{fig:review_scores}
\end{figure}

\paragraph{Scientific slop reads human-written papers the way their reviewers do.}
Figure~\ref{fig:review_scores}(a) compares each system's AI probability
with human review ratings.
Scientific slop shows a clearer \textbf{relationship with review scores}
than the evaluated detectors and automated reviewers.
Its AI probability declines more sharply as ratings rise,
indicating that the score also captures differences in
reviewer-assessed quality.
Figure~\ref{fig:review_scores}(b) examines individual review dimensions.
Only scientific slop consistently assigns higher scores to papers rated
lower across overall rating, soundness, presentation, and contribution,
with statistical support in all four dimensions.
This links scientific slop to \textbf{soundness and contribution}
as well as presentation.
Figure~\ref{fig:review_scores}(c) compares the separation of rejected
and accepted papers across years.
Scientific slop exceeds chance in all nine years, whereas detectors
fall below chance in most comparisons.
The relationship thus extends to \textbf{acceptance decisions}
across changing review scales.
Together, these findings suggest that the weaknesses we define
as scientific slop are already reflected in lower human evaluations.
Scientific slop makes these recurring weaknesses explicit and measurable.
Measurement details are provided in Appendix~\ref{app:measurers_review}.

\subsection{Does \MitigateName Remove the Slop That General Revision Leaves?}
\label{sec:SciSlopHarness_revision_result}

We evaluate whether \MitigateName removes scientific slop that survives
general revision without receiving human anchors as revision targets.
We run \MitigateName and the four revision baselines on every AI-generated
paper in \benchmarkName, revising each paper for three rounds under the
shared protocol of Appendix~\ref{app:baseline_protocol}.
We first compare it with general revision, then examine whether repairs
guided by connections in scientific reasoning approach the reference
levels observed in human papers.

\begin{figure}[t]
\centering
\includegraphics[width=\linewidth]{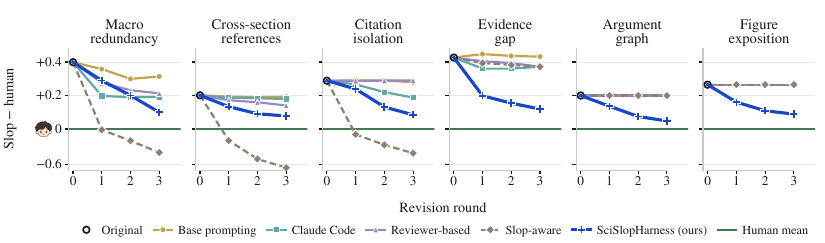}
\caption{Scientific slop across revision rounds for \textcolor{ourscolor}{\MitigateName} and baselines.
Values closer to zero (\textcolor{humancolor}{green}) are closer to matched human means, with round 0 denoting original manuscripts and the scale below zero compressed.}
\label{fig:SciSlopHarness_result}
\end{figure}

\paragraph{\MitigateName reduces the AI–human gap that general revision leaves and slop-aware revision overcorrects.}
Figure~\ref{fig:SciSlopHarness_result} shows that \MitigateName
ends closest to the human mean on all six measures, despite never
receiving these means as revision targets.
For \emph{Macro redundancy}, \emph{Cross-section references}, and
\emph{Citation isolation}, general revision leaves substantial slop,
whereas slop-aware revision pushes scores past the human means
and increasingly away from them, suggesting overcorrection.
\MitigateName reduces the defects left by general revision
while remaining closer to human levels.
Its gains also extend to \emph{Evidence gap}, \emph{Argument graph},
and \emph{Figure exposition}, where the baselines make little
or no progress.
The harness evaluates each defect in the context of the paper's
scientific reasoning.
The skill guides edits that connect claims with supporting evidence
where needed, while review checks whether each edit contributes
to the argument and preserves existing information.
Together, they address defects left by feedback alone while
limiting overcorrection.
Summed over the six measures, the remaining gap to the human means is 63\% smaller than for Claude Code, the closest baseline (Appendix~\ref{app:scislop-gap}).
The approach toward human levels thus supports scientific
reasoning as a criterion for deciding what to repair and what
to preserve, without specifying a target score.

\subsection{Is the Slop Removed, or Only the Score?}
\label{sec:slop_removed_to_score}

We test whether reductions in scientific slop reflect valid repairs rather than lower scores alone.
We compare the revisions of \methodName{} with slop-aware revision to see what produces these reductions, and ablate its components.

\begingroup
\setlength{\intextsep}{6pt}
\begin{table}[t]
\centering
\caption{The same passage before revision, after slop-aware revision, and after \MitigateName, marking \orig{the located slop}, \redt{what a revision lost or left}, and \greent{what it kept or repaired}. Brackets replace citation and reference commands.} 
\captionsetup{skip=4pt,belowskip=0pt}
\scriptsize
\setlength{\tabcolsep}{4pt}
\renewcommand{\arraystretch}{1.05}
\sloppy  %
\makeatletter\def\SOUL@hlpreamble{\setul{0.35ex}{1.75ex}\let\SOUL@stcolor\SOUL@hlcolor\SOUL@stpreamble}\makeatother
\definecolor{hlred}{RGB}{250,214,214}
\definecolor{hlgreen}{RGB}{208,236,212}
\definecolor{hlgray}{RGB}{222,222,222}
\definecolor{markred}{RGB}{190,55,55}
\definecolor{markgreen}{RGB}{50,135,60}
\providecommand{\redt}[1]{{\sethlcolor{hlred}\hl{#1}}}
\providecommand{\greent}[1]{{\sethlcolor{hlgreen}\hl{#1}}}
\providecommand{\rowbad}{\textcolor{markred}{$\times$}}
\providecommand{\rowgood}{\textcolor{markgreen}{$\checkmark$}}
\providecommand{\lightrule}{\noalign{\vskip 1.6pt{\color{black!22}\hrule height 0.3pt}\vskip 1.6pt}}
\providecommand{\dotrule}{\noalign{\vskip 1.6pt{\color{black!38}\hbox to \linewidth{\leaders\hbox to 2.6pt{\hss\rule[0pt]{0.6pt}{0.6pt}\hss}\hfill}}\vskip 1.6pt}}
\providecommand{\orig}[1]{{\sethlcolor{hlgray}\hl{#1}}}
\begin{tabularx}{\linewidth}{@{}l@{\hspace{3pt}}c>{\arraybackslash}X@{}}
\toprule
\multicolumn{3}{@{}>{\arraybackslash}p{\linewidth}@{}}{\textcolor{structurecolor}{\textbf{Macro redundancy}}\quad Shaded sentence repeats the Abstract. \textcolor{markred}{$\times$ Slop-aware drops the citations.} \textcolor{markgreen}{$\checkmark$ \methodName keeps them.}} \\
\lightrule
Original & & \orig{RLVR has emerged as a powerful paradigm\ldots}\ from early work on \orig{CodeRL [1] and RLTF [2]} to recent breakthroughs like \orig{DeepSeek-R1 [3] and DeepSeekMath [4]}\ldots \\
\lightrule
Slop-aware & \rowbad & \ldots\ \redt{Early work in Section [Related Work]} demonstrates\ldots\ from \redt{pioneering applications of unit test signals} to recent breakthroughs like \redt{DeepSeek-R1 and DeepSeekMath}\ldots \\
\dotrule
\methodName & \rowgood & In code generation, unit tests provide a natural verifier\ldots\ from early work on \greent{CodeRL [1] and RLTF [2]} to recent breakthroughs like \greent{DeepSeek-R1 [3] and DeepSeekMath [4]}\ldots \\
\addlinespace[3pt]
\midrule
\multicolumn{3}{@{}>{\arraybackslash}p{\linewidth}@{}}{\textcolor{structurecolor}{\textbf{Cross-section references}}\quad Table [tau] is unreferenced. \textcolor{markred}{$\times$ Slop-aware scatters references.} \textcolor{markgreen}{$\checkmark$ \methodName adds one where shaded.}} \\
\lightrule
Original & & We presented Delta-Prefill Switching (DPS)\ldots\ DPS achieves 21--22\% speedup over greedy decoding\ldots\ \orig{The method is robust to threshold selection} and requires no model modifications. \\
\lightrule
Slop-aware & \rowbad & As positioning against prior work \redt{(Section [Related Work])} shows\ldots\ Delta-Prefill Switching \redt{(Section [Method])} addresses a fundamental tension\ldots\ Our experiments \redt{(Section [Experiments])} demonstrate\ldots\ Sequential performance \redt{(\mbox{Table [sequential]})} shows DPS achieves 21--22\% speedup\ldots\ concurrent evaluation \redt{(\mbox{Table [concurrent]} and \mbox{Figure [concurrent]})} reveals\ldots\ \redt{Threshold sensitivity analysis (Table [tau] and Figure [tau]) confirms robustness across a wide range of parameter values.} \\
\dotrule
\methodName & \rowgood & We presented Delta-Prefill Switching (DPS)\ldots\ DPS achieves 21--22\% speedup over greedy decoding\ldots\ The method is robust to threshold selection \greent{(Table [tau] demonstrates that all $\tau \geq 32$ outperform the greedy baseline)} and requires no model modifications. \\
\bottomrule
\end{tabularx}

\label{tab:revision-cases}
\end{table}
\endgroup

\paragraph{\MitigateName removes the slop where slop-aware revision removes only the score.}
Table~\ref{tab:revision-cases} shows how slop-aware revision and \MitigateName repair the same detected problems differently.
For Macro redundancy, slop-aware revision reaches a zero score by removing the repeated sentence together with four citations, losing the attribution to prior work.
In contrast, \MitigateName removes only the repetition and preserves all four citations, reducing redundancy without losing the information needed for the argument.
For Cross-section references, slop-aware revision approaches zero by adding references to three sections, three tables, and two figures throughout the conclusion.
These references improve the score, but most only tell the reader where to look without connecting the evidence to a claim.
In contrast, \MitigateName adds one reference where the threshold table supports the robustness claim, and states what the table shows, that every $\tau \geq 32$ beats the greedy baseline.
\MitigateName reduces the same slop by preserving necessary information and repairing the missing connections between claims and evidence.
We observe the same distinction for the remaining measures in Appendix~\ref{app:revision-cases-v14}.

\begin{wrapfigure}[9]{r}{0.55\linewidth}
\vspace{-7pt}
\centering
\includegraphics[width=\linewidth]{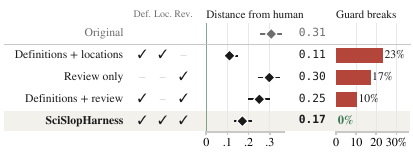}\vspace{-6mm}
\caption{Ablation of \MitigateName.
}
\label{fig:component_summary}
\vspace{-8pt}
\end{wrapfigure}

\paragraph{The full \MitigateName is needed for reliable slop reduction.}
Figure~\ref{fig:component_summary} shows that removing any part of the harness introduces a different failure.
Definitions and locations reduce the distance to 0.11, but without review this comes with guard breaks in 23\% of rounds.
Review alone leaves most slop unrepaired at 0.30 and still breaks a guard in 17\% of rounds.
Definitions with review are safer, but without located instances the distance remains 0.25.
In contrast, the full \MitigateName reduces the distance to 0.17 with no guard breaks.
The ablations show why the components must work together: the harness must identify where to repair, guide the repair, and verify that the resulting edit preserves the scientific record.

\section{Discussion}
\label{sec:conclusion}

\paragraph{Conclusion.}
We introduce \textbf{scientific slop} as measurable breakdowns in the reasoning that connects a paper's parts.
Its measures distinguish AI-generated from human papers and are associated with lower scores across all four ICLR review dimensions.
\MitigateName{} uses the identified defects to guide revisions checked against the paper's own experiment outputs and code.
It brings all six measures closest to human levels among the evaluated revision methods, without receiving human reference targets.
Scientific slop connects detection to weaknesses reflected in human reviews and to repairs supported by scientific records.

\paragraph{Limitations and future directions.} The six scientific slops offer an initial view of how connections within a paper reveal defects and guide repair. Future work can extend this view to how a study’s question, methods, and evidence support its conclusions as a whole. Our evaluation focuses on FARS and Agents4Science 2025, with most papers drawn from computer science. Extending the paired evaluation to new generators and disciplines can test how broadly this view applies.

\section*{AI Use Statement}
LLMs are used in this work both as the subject of study and as components of our experiments. SciSlopBench includes third-party AI-generated papers, our measures use Qwen2.5 models, and SciSlopHarness and its baselines use Claude models as editors and reviewers. 
We also used Claude (including Claude Code) and ChatGPT during implementation and manuscript preparation. They assisted with writing and debugging code for the measurement pipeline, harness, baselines, and analysis scripts, as well as editing the manuscript for grammar, clarity, and concision. In some cases, they were used to produce initial text that was subsequently edited by the authors. The authors reviewed and tested all assisted code and reviewed the final manuscript. All reported results were obtained by running the released code.

We take responsibility for the final content of this work, including text,
claims or artifacts produced with the aid of generative AI.

\section*{Ethics statement}
This work aims to support scientific scrutiny by making failures in reasoning identifiable and open to inspection.
Scientific slop measures give authors and reviewers concrete points to examine in a paper, including weaknesses that fluent prose can obscure.
These findings concern the paper's reasoning and do not establish an author's intent or misconduct.

Correcting a reasoning failure remains valuable even when it weakens a signal of AI authorship.
The criterion for responsible repair is whether the study's records support the change.
\MitigateName{} applies this criterion by checking revisions against the manuscript and experiment records and recording preservation checks for citations and reported numbers.
When no concrete example exists in the records, the revision acknowledges that absence.
This makes the basis for a revision available for inspection alongside its effect on the score.

Our ICLR analysis uses public manuscripts and review information and reports aggregate relationships between measured slop and review outcomes.
The benchmark's data sources and processing procedures are documented to support scrutiny of its construction.

\subsubsection*{Acknowledgments}
This work was supported by 
Institute of Information \& communications Technology Planning \& Evaluation (IITP) grant funded by the Korea government(MSIT) (No.RS-2026-25524173, Ultro-Long-Term Hierarchical Memory and Reasoning Architecture for Next-Generation Omnimodal Agents, 
the Institute of Information \& Communications Technology Planning \& Evaluation(IITP) grant funded by the Korea government(MSIT) (RS-2025-25442338, AI star Fellowship Support Program(Seoul National Univ.)),
the IITP(Institute of Information \& Communications Technology Planning \& Evaluation)-ITRC(Information Technology Research Center) grant funded by the Korea government(Ministry of Science and ICT)(IITP-2025-RS-2024-00437633),
Basic Science Research Program through the National Research Foundation of Korea(NRF) funded by the Ministry of Education(RS-2023-00274280),
National IT Industry Promotion Agency(NIPA) grant funded by the Korea government(MSIT), and
the AI Seoul Tech Research Support Program of the Seoul
Future Foundation.
Gunhee Kim and Dongyeop Kang are the corresponding author.

\appendix

\section{Measurement Specification of the Slop Measures}
\label{app:measurers}

This appendix provides the detailed measurement procedure for each measure in Table~\ref{tab:slop-registry}.

\paragraph{Document views.}
We expand all \verb|\input| files and read each paper as a single \LaTeX{} document.
We define the paper body as the text before the appendix, acknowledgements, or bibliography, whichever comes first, and apply the same rule to both papers in each pair.
We construct two views of this body.
The \emph{structure view}, used by the six scientific slop measures, preserves sections, labels, references, citations, and floats.
The \emph{prose view}, used by the text detectors, removes floats and captions and replaces mathematics, citations, and references with sentinel tokens.
For human papers, we construct these views directly from the \LaTeX{} e-print.
Among the 247 Agents4Science submissions, 18 provide manuscript source.
For the remaining 229, we reconstruct the same \LaTeX{} structure from the PDF, recovering headings, float labels, citations, and references.

\paragraph{Macro redundancy.}
A sentence is recycled when more than half of its tokens sit inside 8-grams that an earlier section already used.
Over the prose view of the abstract and the body sections, for a sentence $u$ of at least eight tokens, with $C(u)$ the tokens of $u$ that lie in an 8-gram first seen in an earlier sentence of a different section,
\begin{equation}
h(u)=\mathbf{1}\!\left[\,|C(u)|/|u|\geq 0.5\,\right].
\end{equation}

\paragraph{Cross-section references.}
Every body section, and every figure, table, equation, algorithm, or theorem labeled inside one, is an object the paper declares.
$R(o)$ collects the sections other than its home that point to $o$ through a reference command, a printed section number, or a named reference, and
\begin{equation}
h(o)=\mathbf{1}\!\left[\,R(o)=\emptyset\,\right].
\end{equation}
Pointing at a table from the section that holds it is neither rewarded nor punished, since only other sections count.
Pointers into the appendix, and a roadmap in the first section, the longest run of forward pointers to later sections, are recorded apart because neither is a trace of one section having read another.

\paragraph{Argument graph.}
Does the context that supports a claim come before the claim?
A language model labels each sentence of the introduction, by a forced choice over the numbered introduction with the majority of three greedy runs, and the key claims are the sentences labeled superiority, prior limitation, or design choice.
The supporting context of $s_{i}$ is the sentence that raises its likelihood most under a second model that is never prompted,
\begin{equation}
\begin{gathered}
\mathrm{pmi}(j\!\to\! i)=\frac{1}{|s_{i}|-1}\sum_{t=2}^{|s_{i}|}\Big[\log P\!\left(s_{i}^{t}\mid s_{j},\,s_{i}^{<t}\right)-\log P\!\left(s_{i}^{t}\mid s_{i}^{<t}\right)\Big],
\\
\pi(i)=\arg\max_{j\neq i}\,\mathrm{pmi}(j\!\to\! i),
\end{gathered}
\label{eq:pmi}
\end{equation}
and the claim is marked when that context comes after it,
\begin{equation}
h(i)=\mathbf{1}\!\left[\,\pi(i)>i\,\right].
\label{eq:ag-score}
\end{equation}

\paragraph{Citation isolation.}
A citing sentence weaves when it cites two works, names a second work outside a citation command, or links two works by a cue from a frozen list, and it is isolated otherwise,
\begin{equation}
h(u)=\mathbf{1}\!\left[\,u\ \text{does not weave}\,\right].
\end{equation}
The score runs over the citing sentences of the introduction and the related work.
Relating a work only to the present paper is not weaving, and fewer than eight citing sentences flags the score weak.

\paragraph{Figure exposition.}
The method figure is read as text.
One vision model transcribes it, with the same prompt on both sides, and six kinds of exposition are checked on the transcript, a notation key, a comparison arm, experimental content, an evaluative mark, a thesis box, and enumerated stages, with $h(k)=1$ when the figure carries kind $k$.
Five kinds are frozen patterns.
Experimental content, whether a number is a setting of the run or a constant of the method, is decided by one reader, on 33 clear and 25 borderline figures of the FARS half and 13 and 9 of the Agents4Science half.
On the FARS half the method figure is the pipeline's overview image, and on the Agents4Science half both sides are located by the same caption gate.

\paragraph{Evidence gap.}
Here the unit is the paper.
It is applicable when its body holds a result table with at least two data rows and four numeric cells, and $h(p)=1$ when no exhibit appears anywhere in it, appendix included.
An exhibit is a display environment carrying consumed or produced material, a caption announcing an example, a case, or a failure, or a long quotation outside the related work, and one exhibit anywhere closes the gap.

\paragraph{Aggregate score.}
The \scislop{} score of a paper averages the item scores of Eq.~\ref{eq:slop-item-score} within each plane and then averages the planes. With $\mathcal{J}_k(p)$ the items of plane $k$ that apply to paper $p$ and $\mathcal{K}(p)$ the planes with at least one such item,
\begin{equation}
S(p)=\frac{1}{|\mathcal{K}(p)|}\sum_{k\in\mathcal{K}(p)}\frac{1}{|\mathcal{J}_k(p)|}\sum_{j\in\mathcal{J}_k(p)}s_j(p).
\label{eq:slop-aggregate}
\end{equation}
This is the \scislop{} row of Tables~\ref{tab:main_results} and~\ref{tab:slop_measures}; the negative of $S(p)$ is used wherever a higher score should favor human authorship.

\paragraph{Models and serving.}
\emph{Argument graph} labels sentences with Qwen2.5-32B-Instruct~\citep{yang2024qwen25} and reads probabilities from Qwen2.5-7B-Instruct, and \emph{Figure exposition} transcribes with Qwen2.5-VL-32B-Instruct~\citep{bai2025qwen25vl}.
Models run on vLLM~\citep{kwon2023vllm} in bfloat16 with greedy decoding, a label or scalar is the majority or median of three runs, every call is cached under a hash of prompt and model, and no measurement calls a paid or closed model.

\section{\benchmarkName{} Construction Details}
\label{app:data_detail}

\benchmarkName{} holds 390 pairs.
Each pair is one AI-generated paper and one human-written paper, its anchor, of the same kind of contribution on the same or a neighboring problem, and every anchor is a paper whose human authorship and peer acceptance can be shown.
The AI papers come from two sources, 143 written end to end by FARS~\citep{tang2026fars} and the 247 submissions to Agents4Science 2025, and the same pairing rule is applied to both.
Tables~\ref{tab:main_results} and~\ref{tab:slop_measures} pool the 390 pairs into one evaluation.
Appendix~\ref{app:pair_rule} gives the pairing rule, Appendix~\ref{app:data_detail_matching} the FARS half, Appendix~\ref{app:a4s_collection} the Agents4Science half and how its PDFs are read, and Appendix~\ref{app:bench_comparison} the relation to existing detection benchmarks.

\subsection{Pairing rule}
\label{app:pair_rule}

A human paper can anchor an AI paper only if it passes four conditions.
\begin{itemize}[leftmargin=2.4em, itemsep=1pt, parsep=0pt, topsep=2pt]
    \item[H1] \textbf{Human authorship.} Its first arXiv version is dated 2024 or earlier, or it is an ICLR 2026 submission whose Pangram AI share is at most 0.05. The date stands in for a check we cannot run on older papers, and Pangram is the only check available for papers written after 2024.
    \item[H2] \textbf{Acceptance.} It was accepted to the main track of a top-tier venue, shown by the venue record, the arXiv comment or journal reference, or OpenReview. Rejected submissions, workshop papers, and preprints do not pass.
    \item[H3] \textbf{Contribution type.} It is the same kind of paper as the AI paper, one of method, framework, benchmark, dataset, analysis, or survey.
    \item[H4] \textbf{Usable source.} Its arXiv e-print expands in document order to a body of at least 2,500 words in at least four sections, with the appendix separable from the body.
\end{itemize}

Within each contribution type, we select the human paper closest to the AI paper in research problem. A language model assigns topic similarity from the title and abstract on four tiers: 3 for the same problem and contribution type, 2 for the same problem, 1 for the same area, and 0 for papers cited only as tools, models, or datasets. We select the highest tier and break ties by experiment-section citation, availability of public review scores, a human-to-AI body-length ratio within 2.5, and finally SPECTER2 cosine similarity~\citep{singh2023scirepeval}. We use cosine only as the final tie-break because candidate similarities are tightly clustered between 0.90 and 0.97.

We search two candidate pools in order. We first consider papers cited by the AI paper, since these are the works against which the paper positions its contribution. If no cited paper satisfies the pairing conditions, we search 2,690 accepted ICLR 2026 papers with an arXiv identifier and Pangram share at most 0.05, retrieve the three nearest by TF-IDF over title and abstract, and apply the same criteria. Assignment is one-to-one within each source, with contested anchors assigned to the closer match. We retain the candidate pool, similarity tier, condition outcomes, and alternative candidates for each pair so that every match can be independently checked.

\subsection{FARS: collection and matching}
\label{app:data_detail_matching}

\paragraph{Source.}
FARS contains 166 papers generated end to end from a research idea, including experiments and manuscript writing. Of these, 165 provide \LaTeX{} source and 160 provide code. Most are method papers (123), followed by analysis (38), framework (4), and benchmark (1).

\paragraph{Matching.}
Of the 166 FARS papers, 161 cite at least one arXiv paper, yielding 1,946 candidates from 1,197 unique papers. Applying H1–H4 leaves candidates for 124 papers. H3 removes the most because many citations are benchmarks or base models rather than papers of the same contribution type. One-to-one assignment then yields 121 pairs from the cited pool, resolving 16 shared-anchor conflicts in favor of the closer match. For the remaining 45 papers we search the widened ICLR 2026 pool, which pairs 23 more; the other 22 (13 analysis, 7 method, 2 framework) have no same-type anchor in either pool and are not paired. One of the 144 pairs is dropped because its FARS paper ships no \LaTeX{} source, leaving the 143 pairs of the benchmark, 120 from the cited pool and 23 from the widened pool (Table~\ref{tab:alignment}).

\begin{table}[!ht]
\centering
\footnotesize
\setlength{\tabcolsep}{4pt}
\renewcommand{\arraystretch}{1.15}
\setlength{\aboverulesep}{0pt}\setlength{\belowrulesep}{0pt}
\begin{tabular}{@{}l r r rrr r@{}}
\toprule
\headrow Stratum & Pairs & Mean cosine & Rank 1 & Rank $\le 3$ & Rank $\le 10$ & Median rank \\
\midrule
\textbf{All pairs} & \textbf{143} & \textbf{0.927} & 37 & 71 & 101 & 4 \\
\hspace{0.6em}Cited pool & 120 & 0.928 & 33 & 61 & 85 & 3 \\
\hspace{0.6em}Widened pool & 23 & 0.919 & 4 & 10 & 16 & 5 \\
\textcolor{black!60}{Cross-pair reference} & \textcolor{black!60}{20,306} & \textcolor{black!60}{0.881} & & & & \\
\bottomrule
\end{tabular}
\caption{
Retrieval audit of the assignment with SPECTER2 similarity between the AI paper and the
human anchor. Rank is the position of the assigned anchor among all 143 anchors for that
AI paper, and the reference row gives the mean cosine over every non-assigned pair.
}
\label{tab:alignment}
\end{table}

\subsection{Agents4Science: collection, matching, and document processing}
\label{app:a4s_collection}

\paragraph{Collection.}
Agents4Science 2025 provides 247 primarily AI-authored submissions, and we include all of them. Most are in computer and data sciences (178), particularly AI and machine learning (154). By contribution type, 171 are method papers, 46 analysis, 19 framework, 6 survey, and 5 benchmark. Acceptance decisions, reviewer scores, and reported AI involvement are retained as metadata but are not used for pairing. Tables~\ref{tab:a4s_primary_topics} and~\ref{tab:a4s_secondary_topics} report the full topic distribution.

\begin{table}[!t]
\centering
\begin{minipage}[t]{0.43\linewidth}\vspace{0pt}
\centering
\footnotesize
\setlength{\tabcolsep}{3.5pt}
\setlength{\aboverulesep}{0pt}\setlength{\belowrulesep}{0pt}
\renewcommand{\arraystretch}{1.08}
\begin{tabular}{@{}lrrr@{}}
\toprule
\headrow Primary topic & Total & Acc. & Rej. \\
\midrule
Computer \& Data Sciences & 178 & 39 & 139 \\
Life \& Health Sciences & 17 & 2 & 15 \\
Social Sciences & 14 & 5 & 9 \\
Natural Sciences & 13 & 1 & 12 \\
Interdisciplinary & 11 & 0 & 11 \\
Engineering \& Technology & 6 & 1 & 5 \\
Mathematics \& Statistics & 4 & 0 & 4 \\
Humanities \& Arts & 2 & 0 & 2 \\
Law, Policy \& Business & 1 & 0 & 1 \\
Not specified & 1 & 0 & 1 \\
\midrule
\textbf{Total} & \textbf{247} & \textbf{48} & \textbf{199} \\
\bottomrule
\end{tabular}
\caption{Primary topics and acceptance status of the 247 Agents4Science 2025 papers listed in the public submission table (Acc./Rej.: accepted/rejected). One rejected submission has no domain annotation.}
\label{tab:a4s_primary_topics}

\end{minipage}\hfill
\begin{minipage}[t]{0.54\linewidth}\vspace{0pt}
\centering
\footnotesize
\setlength{\tabcolsep}{4pt}
\renewcommand{\arraystretch}{1.03}
\setlength{\aboverulesep}{0pt}\setlength{\belowrulesep}{0pt}
\begin{tabularx}{\linewidth}{@{}Xrrr@{}}
\toprule
\headrow Primary / secondary topic & Total & Acc. & Rej. \\
\midrule
\multicolumn{4}{@{}l}{\textit{Computer \& Data Sciences}} \\
\quad AI \& Machine Learning & 154 & 32 & 122 \\
\quad Human-Computer Interaction & 19 & 6 & 13 \\
\quad Data Science & 3 & 1 & 2 \\
\quad Software \& Systems & 2 & 0 & 2 \\
\addlinespace[2pt]
\multicolumn{4}{@{}l}{\textit{Life \& Health Sciences}} \\
\quad Medicine \& Health & 10 & 0 & 10 \\
\quad Biology & 4 & 2 & 2 \\
\quad Agriculture \& Ecology & 2 & 0 & 2 \\
\quad Neuroscience & 1 & 0 & 1 \\
\addlinespace[2pt]
\multicolumn{4}{@{}l}{\textit{Social Sciences}} \\
\quad Economics & 5 & 3 & 2 \\
\quad Psychology & 5 & 1 & 4 \\
\quad Communication \& Media & 2 & 0 & 2 \\
\quad Operations Research & 1 & 1 & 0 \\
\quad Education & 1 & 0 & 1 \\
\addlinespace[2pt]
\multicolumn{4}{@{}l}{\textit{Natural Sciences}} \\
\quad Physics & 9 & 0 & 9 \\
\quad Astronomy & 4 & 1 & 3 \\
\addlinespace[2pt]
\multicolumn{4}{@{}l}{\textit{Interdisciplinary}} \\
\quad Cognitive Science & 5 & 0 & 5 \\
\quad Computational Social Science & 4 & 0 & 4 \\
\quad Science \& Technology Studies & 1 & 0 & 1 \\
\quad Biomedical Engineering & 1 & 0 & 1 \\
\addlinespace[2pt]
\multicolumn{4}{@{}l}{\textit{Engineering \& Technology}} \\
\quad Materials Science & 4 & 1 & 3 \\
\quad Biomedical Engineering & 1 & 0 & 1 \\
\quad Electrical \& Computer Engineering & 1 & 0 & 1 \\
\addlinespace[2pt]
\multicolumn{4}{@{}l}{\textit{Mathematics \& Statistics}} \\
\quad Pure \& Applied Mathematics & 3 & 0 & 3 \\
\quad Operations Research & 1 & 0 & 1 \\
\addlinespace[2pt]
\multicolumn{4}{@{}l}{\textit{Humanities \& Arts}} \\
\quad Philosophy & 1 & 0 & 1 \\
\quad History & 1 & 0 & 1 \\
\addlinespace[2pt]
\multicolumn{4}{@{}l}{\textit{Law, Policy \& Business}} \\
\quad Public Policy & 1 & 0 & 1 \\
\midrule
Not specified & 1 & 0 & 1 \\
\textbf{Total} & \textbf{247} & \textbf{48} & \textbf{199} \\
\bottomrule
\end{tabularx}
\caption{Secondary topics within each primary category for the 247 Agents4Science 2025 papers listed in the public submission table. Rejected counts are the difference between total and accepted counts. Human-Computer Interaction combines two punctuation variants. Biomedical Engineering and Operations Research each appear under two primary categories.}
\label{tab:a4s_secondary_topics}

\end{minipage}
\end{table}

\paragraph{Matching.}
Every one of the 247 submissions is paired with a human anchor that passes H1, H2, and H4. The full pairing rule, which also requires the same contribution type, pairs 185 of them: 66 anchors cited by the submission and 119 from the widened ICLR 2026 pool at tier 2 or higher. The remaining 62 submissions cover topics poorly represented in ICLR, including physics, astronomy, biology, medicine, finance, and human-computer interaction, so for these we keep H1, H2, and H4 fixed and take the closest admissible anchor in a recorded order. Of the 62, 19 share the research problem, 21 share the contribution type and the area, 16 share the area, and 6 are matched to the nearest TF-IDF candidate. The last 22 are thus still paired with an accepted, human-written paper, at the area level or by nearest neighbor, rather than left without a counterpart. Overall, 206 of the 247 anchors match the contribution type and 196 match the research problem. We record the match level for every pair, allowing evaluation on the strict 185 pairs, the 196 problem-matched pairs, or all 247.

\paragraph{Document processing.}
We read human anchors from their arXiv \LaTeX{} source. Among the AI submissions, 18 have \LaTeX{} source consistent with the submitted PDF and are read directly; the remaining 229 are reconstructed from PDF into the \LaTeX{} elements required by our measures. The reconstruction extracts section headings, figure and table captions, citations, cross-references, and result-table cells using PyMuPDF. It does not infer content absent from the PDF. On the 18 submissions available in both formats, reconstruction recovers a median of 20 body headings compared with 21 in the source. We record the processing route for every paper.

\subsection{Comparison with existing benchmarks}
\label{app:bench_comparison}

\newcommand{\cyes}{\textcolor{humancolor}{\ensuremath{\checkmark}}}
\newcommand{\cpart}{\textcolor{revcolor}{\ensuremath{\bigcirc}}}
\newcommand{\cstud}{\textcolor{detcolor}{\ensuremath{\triangle}}}
\newcommand{\cno}{\textcolor{black!45}{--}}
\begin{table}[!ht]
\centering
\scriptsize
\setlength{\tabcolsep}{4pt}
\renewcommand{\arraystretch}{1.2}
\setlength{\aboverulesep}{0pt}\setlength{\belowrulesep}{0pt}
\begin{tabular}{@{}l cccccccc@{}}
\toprule
\headrow Benchmark $\downarrow$\ \ Property $\rightarrow$
& \shortstack{Scientific\\text}
& \shortstack{Whole\\paper}
& \shortstack{Tables and\\figures}
& \shortstack{Released\\code}
& \shortstack{Executed\\research}
& \shortstack{Matched\\human pair}
& \shortstack{Multiple\\genres}
& \shortstack{Adversarial\\variants} \\
\midrule
RAID~\citep{dugan2024raid}      & \cyes & \cno  & \cno & \cno & \cno & \cyes  & \cyes & \cyes  \\
DetectRL~\citep{wu2024detectrl} & \cyes & \cno  & \cno & \cno & \cno & \cpart & \cyes & \cyes  \\
M4~\citep{wang2024m4}           & \cyes & \cno  & \cno & \cno & \cno & \cyes  & \cyes & \cno  \\
MAGE~\citep{li2024mage}         & \cyes & \cno  & \cno & \cno & \cno & \cpart & \cyes & \cno  \\
CHEAT~\citep{yu2023cheat}       & \cyes & \cno  & \cno & \cno & \cno & \cyes  & \cno & \cpart \\
IDMGSP~\citep{mosca2023idmgsp}  & \cyes & \cpart & \cno & \cno & \cno & \cyes  & \cno & \cpart \\
\midrule
\oursrow \textbf{\benchmarkName{} (ours)} & \cyes & \cyes & \cyes & \cyes & \cyes & \cyes & \cno & \cstud \\
\bottomrule
\end{tabular}
\caption{
Comparison with existing machine-generated text benchmarks.
(\cyes) the benchmark instances carry the property, (\cpart) partly, such as a human
text from the same source rather than a matched counterpart, (\cstud) studied in the
paper but not in the benchmark instances, (\cno) absent.
}
\label{tab:bench_comparison}
\end{table}

\benchmarkName{} is designed to evaluate scientific failures that passage-level detection benchmarks cannot represent (Table~\ref{tab:bench_comparison}). Existing benchmarks primarily treat passages or abstracts as independent text instances. In contrast, \benchmarkName{} preserves complete papers and pairs each AI paper with a human paper addressing the same research problem. This makes relations across sections, claims, citations, figures, tables, and evidence directly measurable rather than reducing detection to local writing patterns. The FARS half further includes the code that produced the reported results, enabling checks between manuscript claims and experimental records. While existing benchmarks provide broader coverage of genres and adversarial rewrites, \benchmarkName{} provides the paper-level structure and matched scientific context required to evaluate scientific slop.

\clearpage
\section{\MitigateName{} Implementation Details}
\label{app:scislop-harness}

This appendix gives the implementation of \MitigateName{} used for Figure~\ref{fig:SciSlopHarness_result}.
Table~\ref{tab:scislop-config} lists the configuration, and the following subsections describe each stage in the form of the harness summaries of \citet{lee2026meta}.
Figure~\ref{fig:SciSlopHarness_result} reports every AI-generated paper in \benchmarkName{}, each revised for three rounds, and the baselines of Appendix~\ref{app:baseline_detail} are run on the same papers.

\subsection{Configuration}
\label{app:scislop-config}

The editor, the six definitions, and their measurers stay fixed across all conditions, and so does the reviewer.
A round consists of one editor pass, one review pass, and one measurement.

\begin{table}[!ht]
\centering\small
\renewcommand{\arraystretch}{1.1}
\setlength{\aboverulesep}{0pt}\setlength{\belowrulesep}{0pt}
\begin{tabularx}{\linewidth}{@{}>{\raggedright\arraybackslash}p{0.25\linewidth}>{\raggedright\arraybackslash}X@{}}
\toprule
\headrow Setting & Configuration used for Figure~\ref{fig:SciSlopHarness_result} \\
\midrule
Editor & \texttt{claude-haiku-4-5-20251001} \\
Reviewer & \texttt{claude-sonnet-5}, no tools \\
Retirement call & \texttt{claude-haiku-4-5-20251001} \\
Skill & \texttt{SciSlop\_v0.5.md}, staged as \texttt{SciSlop.md} \\
Budget & At most three rounds \\
Measured patterns & All six \\
Figure editing & Located expository phrases erased from the raster method figure by inpainting. The edit is a reviewed change. \\
\bottomrule
\end{tabularx}
\caption{Configuration of the default revision procedure.}
\label{tab:scislop-config}
\end{table}

\subsection{Localized revision}
\label{app:scislop-skill}
\label{app:scislop-editor}

\paragraph{Overview.}
The editor receives two files.
\texttt{SciSlop.md} says how each pattern is repaired, and \texttt{SLOP\_FINDINGS.md} says where the remaining instances are.
It revises one source file per call.
Table~\ref{tab:scislop-repairs} lists the repair direction of each pattern.

\begin{itemize}[leftmargin=1.4em,itemsep=2pt,topsep=3pt]
\item \textbf{Skill file.} \texttt{SciSlop.md} holds the six definitions, each with a repair direction and the format of its located instances. Its global rules are short. Do not fabricate experiments, numbers, citations, or examples. Do not add or remove cited works and do not change reported numbers. Keep the method, the experiments, and the results as they are. Make the smallest edit that removes the pattern. Treat listed locations as candidates and leave one alone when the repair would add text the argument does not need. The file has no score formula and no target.
\item \textbf{Located feedback.} \texttt{SLOP\_FINDINGS.md} lists the remaining instances by pattern, up to 8 per pattern with the count of the rest. From round 2 onward, the previous round's review notes are appended: the reverted changes with the reviewer's reasons, and the retired instances.
\item \textbf{Editor call.} For each of \texttt{main.tex} and \texttt{sections/*.tex}, one call sends \texttt{SciSlop.md}, \texttt{SLOP\_FINDINGS.md}, the other files as read-only context, and the file to edit. The editor edits only its own file. It receives no human paper, no score, and no formula.
\item \textbf{Edit blocks.} The call returns edit blocks, each quoting an original passage and giving its replacement, or the word \texttt{NOCHANGE}. A block is applied only when the quoted passage occurs exactly once in the file. An edited file is accepted only if it keeps the file's structural commands and its length stays within 0.5 to 2.5 times the original.
\item \textbf{Figure editing.} The method figure is a raster image, so the editor lists the expository phrases to erase and a deterministic step erases them. The controller keeps only phrases the measurer had classified as expository. A vision-language model returns a bounding box for each phrase and an inpainting model fills the box from its surroundings. An edit report records the erased phrases, the boxes, and the pixels changed outside them.
\item \textbf{Records.} Before each round, the controller copies the project's own experiment records from \texttt{exp/EXPERIMENT\_RESULTS} into \texttt{materials/}. Once per paper, a specimen search checks whether these records hold a displayable instance and must quote it verbatim from a named file.
\end{itemize}

\begin{table}[!ht]
\centering\small
\renewcommand{\arraystretch}{1.08}
\setlength{\aboverulesep}{0pt}\setlength{\belowrulesep}{0pt}
\begin{tabularx}{\linewidth}{@{}>{\raggedright\arraybackslash}p{0.21\linewidth}>{\raggedright\arraybackslash}X>{\raggedright\arraybackslash}p{0.21\linewidth}@{}}
\toprule
\headrow Pattern & Repair direction & Reported location \\
\midrule
Cross-section references & Refer to an object from one sentence that uses it, placed where the result is interpreted or the method is applied. Leave genuinely local objects unchanged. Add no sentence that only points. & Label, object kind, home section \\
Macro redundancy & Rewrite the recycled sentence so that it adds a detail, consequence, or qualification its section needs, or remove it. Do not paraphrase to disguise repetition. & Section, sentence, source section \\
Argument graph & Move the context that leads up to a shallow claim ahead of it, or move the claim after that context. Do not add, delete, or weaken claims. & Claim sentence and the supporting sentence that currently follows it \\
Citation isolation & State how a cited work relates to another work already cited, in one sentence citing both, using only relations the manuscript already supports. & Section and citing sentence \\
Figure exposition & Remove the expository elements and state them once in the text if needed. Keep every component and connection. The editor lists the phrases to erase and a deterministic step erases them. & Figure, element type, transcribed element text \\
Evidence gap & Never invent an instance. \textbf{(a)}~If the experiment records hold a specimen, display it verbatim with its source file. \textbf{(b)}~If they hold none, say so where the aggregate result is interpreted: no individual case was retained or inspected. & One manuscript-level observation: silent, acknowledged, or exhibited \\

\bottomrule
\end{tabularx}
\caption{Repair directions and location formats of \texttt{SciSlop\_v0.5.md}. All six entries supply feedback and enter the stopping rule.}
\label{tab:scislop-repairs}
\end{table}

\subsection{Review against scientific records}
\label{app:scislop-reviewer}

\paragraph{Overview.}
The reviewer reads every change of the round as a before-and-after pair and keeps or reverts each one.
It checks the change against the original manuscript and the project records.

\begin{itemize}[leftmargin=1.4em,itemsep=2pt,topsep=3pt]
\item \textbf{Changes.} The controller aligns $M_{r-1}$ and $\widetilde{M}_r$ paragraph by paragraph. Each differing paragraph becomes a numbered change with its before-and-after text. A figure edit becomes one further change described by its edit report.
\item \textbf{Review call.} Changes are reviewed three at a time, in parallel. Each call receives the rules below, its three changes, the full revised and original text of the files they touch, and the lines of the records that contain any number used in the revised text. It returns keep or revert, a category, and a reason quoting the decisive words for each change.
\item \textbf{Revert rules.} The revised text says something that neither the manuscript nor the records support. A reported number is changed, rounded, dropped, or introduced without an exact match. A reference, section pointer, or citation is added to a sentence that does not use it. Two cited works are joined by a relation the manuscript never stated. A figure, table, label, citation command, or result is removed. A figure edit erased a phrase the check did not list or a component of the mechanism. A new exhibit is not an exact copy of a record with its source named. Citing sentences are merged without a supported relation.
\item \textbf{Keep rules.} The change repairs a listed pattern with material already in the manuscript or records, in a sentence that uses what it refers to, and leaves every number and cited work as it was. A sentence stating that no individual case was inspected and that results rest on aggregate scores is kept when the records hold no specimen.
\item \textbf{Application.} A change mixing supported and unsupported edits is reverted in full, with the reason naming the unsupported part. Only an explicit keep retains a change. Everything else is restored from $M_{r-1}$.
\item \textbf{Retirement.} A separate call proposes instances to leave alone for the rest of the procedure. It must look for a home for the object in every other section and name the sections it checked. The controller applies retirement only to section instances and to the evidence-gap observation. Retired instances leave the feedback but stay in the measurements.
\end{itemize}

\subsection{Measurement, preservation, and stopping}
\label{app:scislop-protocol}

\begin{itemize}[leftmargin=1.4em,itemsep=2pt,topsep=3pt]
\item \textbf{Measurement after review.} The retained manuscript is measured after review, so reverted edits cannot contribute to the reported reduction. The four deterministic measurers run every round. The argument graph is remeasured whenever the introduction changed, with the same labeling server and estimator as the benchmark. The figure exposition is rescored from the transcript of the current image. A manuscript whose measured evidence gap is 1.0 and whose text states that no individual case was retained or inspected and that the claims rest on aggregate scores is recorded as an acknowledged gap of 0.5. The matched human values are the unmodified measurer values.
\item \textbf{Preservation checks.} Table~\ref{tab:scislop-guards} lists the checks computed against $M_0$ after every round. A hard violation disqualifies the round as successful mitigation. Audit flags mark changes that may be legitimate but require inspection, such as a removed recycled sentence that also removed a repeated number. Numbers are compared as tokens after removing comments and the arguments of citation, reference, label, input, and graphics commands. 
\item \textbf{Stopping.} The procedure stops when no instance remains after retirement, when the editor changes nothing, when review restores the whole round twice in a row, or after three rounds. The last retained manuscript is returned with its per-round trees, measurements, verdicts, retirements, edit reports, and preservation records.
\end{itemize}

\begin{table}[!ht]
\centering\small
\renewcommand{\arraystretch}{1.12}
\setlength{\aboverulesep}{0pt}\setlength{\belowrulesep}{0pt}
\newcommand{\hardc}{\textcolor{linkred}{\textbf{Hard}}}
\newcommand{\auditc}{\textcolor{revcolor}{Audit}}
\begin{tabularx}{\linewidth}{@{}>{\raggedright\arraybackslash}p{0.23\linewidth}>{\raggedright\arraybackslash}Xl@{}}
\toprule
\headrow Check & Condition to report relative to $M_0$ & Class \\
\midrule
Cited works & Addition or removal of a cited bibliography key & \hardc \\
Dangling citations & A newly unresolved citation key & \hardc \\
Reference targets & A reference to an absent label & \hardc \\
Numeric values & An original body numeric token is absent & \hardc \\
Body length & Revised-to-original word ratio outside $[0.8,1.25]$ & \hardc \\
Labels & An original label is removed & \auditc \\
New numbers & A numeric token is added & \auditc \\
Numeric mentions & A token originally occurring at least twice occurs fewer times & \auditc \\
Denominators & A deterministic pattern's score and denominator both decrease & \auditc \\
Evidence gap & An originally silent gap is measured as acknowledged or exhibited & \auditc \\
\bottomrule
\end{tabularx}
\caption{Preservation checks recorded after every round. \textcolor{linkred}{\textbf{Hard}} constraints define admissible mitigation. \textcolor{revcolor}{Audit} flags require examination of the underlying change.}
\label{tab:scislop-guards}
\end{table}

\subsection{Distance to human means}
\label{app:scislop-gap}

\begin{table}[!ht]
\centering\scriptsize
\renewcommand{\arraystretch}{1.12}
\setlength{\aboverulesep}{0pt}\setlength{\belowrulesep}{0pt}
\setlength{\tabcolsep}{2pt}
\begin{tabular}{@{}l r rrrr >{\columncolor{oursbg}}r@{}}
\toprule
\headrow Measure & Original & Base prompting & Claude Code & Reviewer-based & Slop-aware & \MitigateName \\
\midrule
Macro redundancy & $+$0.399 & $+$0.314 & $+$0.191 & $+$0.213 & $-$0.400 & \textbf{$\boldsymbol{+}$0.100} \\
Cross-section refs. & $+$0.201 & $+$0.193 & $+$0.180 & $+$0.141 & $-$0.660 & \textbf{$\boldsymbol{+}$0.078} \\
Argument graph & $+$0.200 & $+$0.200 & $+$0.200 & $+$0.200 & $+$0.200 & \textbf{$\boldsymbol{+}$0.048} \\
Citation isolation & $+$0.289 & $+$0.279 & $+$0.188 & $+$0.290 & $-$0.411 & \textbf{$\boldsymbol{+}$0.085} \\
Figure exposition & $+$0.265 & $+$0.265 & $+$0.265 & $+$0.265 & $+$0.265 & \textbf{$\boldsymbol{+}$0.089} \\
Evidence gap & $+$0.427 & $+$0.432 & $+$0.372 & $+$0.372 & $+$0.372 & \textbf{$\boldsymbol{+}$0.119} \\
\midrule
Sum of absolute distances & 1.780 & 1.682 & 1.395 & 1.480 & 2.307 & \textbf{0.519} \\
\bottomrule
\end{tabular}
\caption{Distance from the human mean after the third revision round (revised mean $-$ human mean) on the all papers of Figure~\ref{fig:SciSlopHarness_result}. Bold marks the smallest distance per row.}
\label{tab:harness_gap}
\end{table}

Table~\ref{tab:harness_gap} gives the values behind the 63\% reduction reported in Section~\ref{sec:SciSlopHarness_revision_result}.
For each measure, the distance is the mean score of the revised papers after the third round minus the matched human mean, the quantity plotted in Figure~\ref{fig:SciSlopHarness_result}.
The gap of a method is the sum of its absolute distances over the six measures.
Claude Code has the smallest gap among the four baselines, so the reduction is computed against it.
\MitigateName{} closes the gap from 1.395 to 0.519, a reduction of 63\%, and ends closest to the human mean on every measure.
Slop-aware revision passes below the human mean on Macro redundancy, Cross-section references, and Citation isolation, so its gap grows over rounds although its scores fall.

\clearpage
\section{Revision Baseline Details}
\label{app:baseline_detail}

This appendix specifies the four revision baselines compared with \MitigateName{} in Section~\ref{sec:exp_dis}: base prompting, Claude Code, reviewer-based refinement, and slop-aware revision.
The baselines differ only in the interface through which the editor acts and in the instruction it receives. The editor model, the manuscript state, the round protocol, and the measurement are shared.
Appendix~\ref{app:scislop-harness} specifies \MitigateName{} itself.

\newtcolorbox{baselineprompt}[1]{%
  enhanced,breakable,colback=black!2,colframe=black!25,
  colbacktitle=black!7,coltitle=black,boxrule=0.4pt,arc=1.5mm,
  title={#1},fonttitle=\small\sffamily\bfseries,
  fontupper=\small,fontlower=\small,
  left=9pt,right=9pt,top=6pt,bottom=6pt,
  before skip=10pt,after skip=10pt,
  before upper={\raggedright\setlength{\parskip}{3pt}},
  before lower={\raggedright\setlength{\parskip}{3pt}}}
\newcommand{\baselineagentsystem}{You are revising a LaTeX research paper. Source files (main.tex, sections/*.tex, math\_commands.tex, a .bib file) are in your working directory. Use Read/Edit/Write to modify the .tex files in place. Do not create files outside the paper directory. When done, finish your turn; do not run latex and do not summarize.}
\newcommand{\baselinenofabrication}{Do not fabricate experiments, numbers, or citations; if evidence is missing, narrow the claim instead. Do not add citations that are not already in the paper, and do not change reported numbers.}
\newcommand{\baselinegeneric}{Revise this paper so that it is a better research paper: improve its clarity, organization, argumentation, and presentation, and make sure every part of the paper does its job for the reader. Keep the method, the experiments, and the reported results as they are. \baselinenofabrication}

\subsection{Shared protocol}
\label{app:baseline_protocol}

\paragraph{Editor and rounds.}
Every baseline uses \texttt{claude-haiku-4-5-20251001} as the editor, the same model as the default \MitigateName{} editor in Table~\ref{tab:scislop-config}.
Each paper is revised for three rounds.
Round $r$ starts from a copy of the manuscript retained after round $r-1$, and round 1 starts from the original AI-generated manuscript.
The copied tree contains the root \texttt{.tex}, \texttt{.bib}, \texttt{.bst}, and \texttt{.sty} files and \texttt{sections/*.tex}. Figures and compiled PDFs are not copied, because the editor cannot read them and the measurers do not use them.
Each round issues one editor call, which for agent conditions is one agent session that may contain several tool interactions.

\paragraph{Information given to the editor.}
No baseline receives human papers, matched human reference values, slop scores, score formulas, or thresholds.
All four baselines are instructed not to fabricate experiments, numbers, or citations.
The boxes below show each baseline's system and user messages, including its exact constraints.
Slop-aware revision additionally receives the pattern definitions and located instances described in Appendix~\ref{app:baseline_slopaware}.

\paragraph{Execution record.}
If the editor makes no changes, we retry once. If the retry also makes no changes, we exclude that round.
We save each prompt and revision and evaluate revisions with the same metrics.
We also check for content loss and citations missing from the bibliography.

\subsection{Base prompting}
\label{app:baseline_base}

Base prompting is a plain text condition without tools.
The editor receives the complete \LaTeX{} source of the paper inside the prompt and returns the complete revised source of every file inside file markers.
The user instruction is the generic improvement instruction followed by the format specification.

\begin{baselineprompt}{Base prompting}
\textbf{System.} You are an expert editor of machine-learning research papers written in LaTeX. You return revised LaTeX source only.
\tcblower
\textbf{User.} \baselinegeneric\par
This is revision round $\langle r\rangle$. The complete LaTeX source of the paper follows, file by file. Return the revised source of every file, in the same order, using exactly this format and nothing else:\par
\texttt{<<<FILE \{path\}>>>}\par
(the complete revised content of that file)\par
\texttt{<<<END \{path\}>>>}\par
Return all $\langle n\rangle$ files ($\langle$paths$\rangle$). Keep every \texttt{\textbackslash input}, \texttt{\textbackslash label}, \texttt{\textbackslash cite}, \texttt{\textbackslash ref}, environment and preamble line that a file already has, unless the revision moves or removes prose; never remove \texttt{\textbackslash begin\{document\}}, \texttt{\textbackslash end\{document\}}, \texttt{\textbackslash documentclass}, or \texttt{\textbackslash input} lines. Do not abbreviate or elide any part of a file.\par
$\langle$source of each file, preceded by \texttt{===== FILE \{path\} =====}$\rangle$
\end{baselineprompt}

A returned file replaces the original only if it is present in the output, its length lies between 0.5 and 2.5 times the original, it preserves every \texttt{\textbackslash input}, \texttt{\textbackslash documentclass}, \texttt{\textbackslash begin\{document\}}, and \texttt{\textbackslash end\{document\}} token of the original, and it is not identical to the original.
A file that fails any check keeps its previous content, and the reason is logged.
These checks protect the compilability of the tree and do not inspect the prose.

\subsection{Claude Code}
\label{app:baseline_code}

Claude Code~\citep{anthropic2025claudecode} runs with its stock system prompt in the paper directory, with the same editor model as the other conditions.
It runs in restricted mode, which removes shell and code execution and ignores user configuration, so the available tools are file reading, searching, and editing.
The instruction is the generic improvement instruction of base prompting with one added sentence.

\begin{baselineprompt}{Claude Code}
\textbf{System.} \textit{The stock Claude Code system prompt is used without a custom replacement.}
\tcblower
\textbf{User.} \baselinegeneric\par
Make the changes directly by editing the .tex files in the current directory; do not just give advice.
\end{baselineprompt}

\subsection{Reviewer-based refinement}
\label{app:baseline_review}

Reviewer-based refinement alternates an automated review with an editing session.
At each round, Qwen2.5-32B-Instruct~\citep{yang2024qwen25} reads the current manuscript source and returns a structured review. The review is regenerated from the current manuscript each round, so later rounds respond to the revised paper rather than to the original.
The source is passed as text truncated to 50{,}000 characters. If the model returns no valid JSON, the call is repeated at 24{,}000 and then 14{,}000 characters.

\begin{baselineprompt}{Reviewer-based refinement / Review generation}
\textbf{System.} \textit{No separate system message is supplied.}
\tcblower
\textbf{User.} You are a knowledgeable, critical, but fair reviewer for a top machine-learning conference. Read the paper draft below and write a review of its overall quality: the significance of the problem, the soundness of the approach, the strength of the evidence, and the clarity of the writing.\par
Respond with ONLY compact JSON in exactly this format:\par
\texttt{\{"summary": "<2-4 sentence summary of the paper and its contributions>",}\par
\texttt{ "strengths": ["<strength 1>", "<strength 2>", "<strength 3>"],}\par
\texttt{ "weaknesses": ["<weakness 1>", "<weakness 2>", "<weakness 3>"],}\par
\texttt{ "questions": ["<question 1>", "<question 2>", "<question 3>"],}\par
\texttt{ "rating": <integer 1-10, where 1 = trivial or wrong and 10 = seminal>\}}\par
PAPER (LaTeX source, possibly truncated):\par
$\langle$manuscript source$\rangle$
\end{baselineprompt}

The editor is an agent with a neutral system prompt and with read, edit, and write tools. Web search is denied and shell commands are not permitted.
The review prompt and the editing instruction are general. Neither mentions \scislop, the six patterns, or any measured quantity.

\begin{baselineprompt}{Reviewer-based refinement / Manuscript editing}
\textbf{System.} \baselineagentsystem
\tcblower
\textbf{User.} A reviewer left the following review of this paper. Revise the paper to address the review.\par
$\langle$review JSON$\rangle$\par
Do not fabricate experiments, numbers, or citations; if evidence is missing, narrow the claim instead.
\end{baselineprompt}

\subsection{Slop-aware revision}
\label{app:baseline_slopaware}

Slop-aware revision gives the editor a list of detected slop problems and instructions for fixing them in place of a general review.
The editor interface, system prompt, tools, and budget are the same as in reviewer-based refinement.
At the start of each round, we check the current manuscript and prepare one prompt containing the detected problems.
For each problem, the prompt gives its name, definition, repair instructions, and locations in the manuscript.
The definitions and instructions follow the measurer specifications.
We check the first four patterns of Table~\ref{tab:baseline_slop_items}, adding \emph{Argument graph} and \emph{Figure exposition} for the comparison with \MitigateName{}.
The prompt lists up to 12 instances per pattern and reports the number of additional instances.
For \emph{Evidence gap}, it simply states that the manuscript contains no concrete example.
Only detected problems are included, and a round is skipped if no problems are found.
The editor receives no scoring formulas or target scores.
Table~\ref{tab:baseline_slop_items} gives the definition and repair text of each pattern.

\begin{baselineprompt}{Slop-aware revision}
\textbf{System.} \baselineagentsystem
\tcblower
\textbf{User.} An automated check of this manuscript found the following issues. Revise the paper to address them. \baselinenofabrication{} Edit the .tex files in place.\par
This is revision round $\langle r\rangle$.\par
\#\# Issue 1: $\langle$pattern name$\rangle$\par
$\langle$definition$\rangle$\par
What to do: $\langle$repair direction$\rangle$\par
Findings ($\langle k\rangle$ total; $\langle$unit label$\rangle$):\par
- $\langle$located unit$\rangle$ \quad(up to 12 lines)\par
- (and $\langle k-12\rangle$ more of the same kind; treat them the same way)\par
\#\# Issue 2: \ldots
\end{baselineprompt}

\begin{table}[!ht]
\centering\footnotesize
\renewcommand{\arraystretch}{1.1}
\setlength{\tabcolsep}{4pt}
\setlength{\aboverulesep}{0pt}\setlength{\belowrulesep}{0pt}
\begin{tabularx}{\linewidth}{@{}>{\raggedright\arraybackslash}p{0.15\linewidth}>{\raggedright\arraybackslash}X>{\raggedright\arraybackslash}X@{}}
\toprule
\headrow Pattern & Definition & What to do \\
\midrule
Cross-section references &
A paper declares objects: its body sections and the labelled figures, tables, equations, and algorithms inside them. An object is unused when no section other than the one that contains it ever refers to it, so the rest of the paper never builds on it. &
Where the paper's argument actually relies on an object, refer to it from the section that relies on it (for example, discuss the result of a table in the section whose claim it supports, and point back to the method's equation where the experiments use it). Do not add pointers that the surrounding text does not use. \\
\addlinespace
Macro redundancy &
A sentence is recycled when at least half of it repeats, almost word for word, text that an earlier, different section already contained. &
Rewrite each recycled sentence so that it adds what its own section needs (a new detail, a consequence, a qualification), or remove it if the section does not need it. Do not paraphrase merely to disguise the repetition. \\
\addlinespace
Citation isolation &
A citing sentence is isolated when it cites a single prior work and neither groups it with another work, nor mentions another work elsewhere in the sentence, nor states a relation between two works. A related-work section made of isolated sentences lists prior work instead of positioning it. &
Where the paper's positioning depends on a cited work, state how it relates to other works already cited in the paper: a shared limitation, a difference in assumption or method, or what this paper takes from each. Only use works already cited in the paper; do not add new citations. \\
\addlinespace
Evidence gap &
The paper reports aggregate results but never displays a single concrete instance of what it counts: an example input and output, a case, a failure, a worked example, or a quoted specimen. &
If the paper's own materials in this directory contain such an instance (for example a prompt, a generated output, a case in the appendix), display one where the reader needs it and refer to it from the text. If no instance is available in the materials, do not invent one; instead say explicitly, in the relevant section, that no instance is shown. \\
\addlinespace[2pt]
Argument graph &
A key claim of the Introduction is a sentence asserting superiority, a prior limitation, or a design choice. The claim's strongest contextual cue is the Introduction sentence that most raises the claim's plausibility. The claim is declared rather than argued when that cue occurs after the claim, so the Introduction states the conclusion before building toward it. &
Reorder so that the context precedes the claim it supports. Move the limitation of prior work, the observation, or the design constraint ahead of the sentence that asserts the contribution, or move the assertion down to follow it. Do not delete claims and do not add new claims. Do not weaken a claim to avoid arguing for it. \\
\addlinespace
Figure exposition &
A method diagram should depict the mechanism. Its components and how they connect. A box, panel, or caption element is expository when it carries material that belongs to the text rather than the mechanism. Experimental settings, dataset names, numeric results, interpretive claims, a legend that restates the prose, or a step-by-step narration of the pipeline. &
Remove the expository material from the method diagram and, where the reader needs it, state it once in the text near the figure. Keep only components of the method and their connections. Remove exactly the elements listed as expository and keep every other element and every arrow the image has. Do not invent components the figure does not show, do not simplify the mechanism to make the drawing easier, and do not drop the figure. \\
\bottomrule
\end{tabularx}
\caption{Definitions and repair instructions for slop-aware revision.
The first four rows are used in the four-pattern setting.}
\label{tab:baseline_slop_items}
\end{table}

\clearpage
\section{ICLR Data and Measurement}
\label{app:measurers_review}

Figure~\ref{fig:review_scores} uses two corpora of human-written ICLR submissions, 1,177 papers from 2017 to 2025 and 691 from ICLR 2026, scored by every system of Table~\ref{tab:main_results} with the same scripts, settings, and document views.
The measurers and the calibration were fixed on \benchmarkName{}, and every statistic is computed within a year or within ICLR 2026, because the ICLR rating scale changed over the years.

\paragraph{Corpora.}
ICLR is the one venue where every submission, accepted or rejected, carries public review scores and a decision, and its arXiv \LaTeX{} e-prints give the source that the measures read.
The multi-year corpus therefore holds ICLR 2017 to 2025 submissions with both, nine years that span the arrival of LLM writing.
It was collected in strata of year by integer rating and of year by award, so that every year holds papers from rating 2 to rating 9 and its orals; at the ends of the scale the strata take every paper that exists.
Of 1,381 decided papers, 204 fail the extraction rules below, leaving 1,177, 590 rejected, 434 accepted, and 153 oral, with 78 to 240 papers a year.
The ICLR 2026 corpus serves the review dimensions.
It is the one year with a soundness, presentation, and contribution score for every paper on one rating scale, and the one year with an independent commercial AI judgment of the same full text, an analysis that Pangram published for every ICLR 2026 submission.
Its 691 submissions with an arXiv e-print were sampled over six bands of that AI share crossed with the decision, so that the high-AI tail, mostly rejected and rarely on arXiv, is present; 560 are accepted and 131 rejected.

\paragraph{Exclusion.}
A paper is removed only when its source cannot be measured: a missing \verb|\input| file, a main file that is a supplement, a failed section parse, fewer than three body sections, fewer than five declared objects, a prose view under 1,500 words, or any measure below its minimum denominator (Appendix~\ref{app:measurers}).
ICLR 2026 adds fewer than three reviews and a body length outside the 1st to 99th percentile.
No rule reads a score or a rating, and a removed paper is removed for every system.

\paragraph{Systems and scores.}
Detectors read the prose view, reviewers the body source, and the measures the full source.
Every score is oriented so that a higher value means more AI-like, which reverses Binoculars and the reviewers' Overall rating.
The detectors score all 1,177 papers; the two reviewer models score a subset drawn by rating quintile within each year and filled at random by year and decision, 287 papers for CycleReviewer and 282 for AI Scientist.
The \scislop{} line is the aggregate of Eq.~\ref{eq:slop-aggregate} (Appendix~\ref{app:measurers}) scored on every paper of both corpora.
\emph{Argument graph} was not run on ICLR 2026, and \emph{Evidence gap} and \emph{Figure exposition} apply to 869 and 662 of the multi-year papers.

\paragraph{Panel (a).}
Each system is calibrated once on the FARS pairs of \benchmarkName{} by a logistic regression of the AI label on its raw score, and the fitted curve is applied unchanged to the ICLR papers, giving the probability the system assigns to AI authorship.
Each probability is adjusted for year by subtracting the mean of its own year and adding back the mean over all years, so the axis stays on the probability scale; a paper sits at its rounded mean rating.
Levels 2 to 9 hold 31 to 217 papers each; levels 1 and 10, with 11 and 1 papers, are not drawn.

\paragraph{Panel (b).}
A cell is the Spearman correlation between a system's oriented score and the mean reviewer score on one dimension of ICLR 2026, over the papers the system scored, 686 for the detectors, 691 for Pangram, 680 for \scislop{}, and 338 and 285 for CycleReviewer and AI Scientist.

\paragraph{Panel (c).}
Within each year, the value is the probability that a rejected paper scores more AI-like than an accepted or oral paper of the same year.
A system is drawn when at least 15 papers stand on each side in at least five years; the reviewer models meet this in three years and are not drawn.

\clearpage
\section{Additional Matched Revision Cases}
\label{app:revision-cases-v14}

Section~\ref{sec:slop_removed_to_score} compared slop-aware revision and \MitigateName on the same passages for two measures, Macro redundancy and Cross-section references, and showed that the two lower the score by different means, one by deleting or scattering information and the other by repairing the located passage.
This appendix extends that comparison to the remaining four measures, Citation isolation, Evidence gap, Argument graph, and Figure exposition.

\paragraph{The same distinction holds for the remaining four measures.}
Table~\ref{tab:revision-cases-more} repeats the format of Table~\ref{tab:revision-cases} for the four measures it does not cover and shows the same pattern. Slop-aware revision lowers the score by leaving, deleting, or inventing content, whereas \MitigateName repairs the located passage itself.
\begin{itemize}[leftmargin=1.4em,itemsep=2pt,topsep=3pt]
\item \textbf{Citation isolation} shows whether a repair relates the cited work to anything else.
Slop-aware revision rewrites the CALM sentence but still describes CALM alone, so the located sentence stays isolated; its lower manuscript score comes from changes elsewhere in the paper.
\MitigateName keeps the sentence and adds the contrast with InfMem, which stops at the chunk level and reports a 3.3--5.1$\times$ speedup.
The isolation is removed by adding a relation, not by rewording the sentence.
\item \textbf{Evidence gap} shows what a revision does when the record holds no concrete instance.
Slop-aware revision reaches a zero score by inserting a worked example whose question, responses, and numbers appear nowhere in the manuscript or the project records.
\MitigateName states that the claims rest on aggregate results only, and its review reverts the one example the editor inserted as beyond the record.
The gap stays acknowledged at $0.5$ instead of being filled with an invented instance.
\item \textbf{Argument graph} shows whether a claim follows the method it depends on.
Slop-aware revision leaves the design claim ahead of the method and adds a section pointer instead.
\MitigateName moves the claim directly after the method description without deleting either sentence, and the manuscript score falls to zero.
The repair reorders existing content rather than adding text.
\item \textbf{Figure exposition} shows whether a revision can reach text inside an image.
Slop-aware revision edits source text only, so all eight phrases that restate the paper remain in the diagram and the score stays at its original value.
\MitigateName erases six of the eight phrases while leaving the components and arrows intact.
This comparison is not under equal tools, since the baseline has no image editor, but it shows how far the harness's image edits reach.
\end{itemize}

\begin{table}[!ht]
\centering
\captionsetup{skip=4pt}
\scriptsize
\setlength{\tabcolsep}{4pt}
\renewcommand{\arraystretch}{1.05}
\sloppy  %
\makeatletter\def\SOUL@hlpreamble{\setul{0.35ex}{1.75ex}\let\SOUL@stcolor\SOUL@hlcolor\SOUL@stpreamble}\makeatother
\definecolor{hlred}{RGB}{250,214,214}
\definecolor{hlgreen}{RGB}{208,236,212}
\definecolor{hlgray}{RGB}{222,222,222}
\definecolor{markred}{RGB}{190,55,55}
\definecolor{markgreen}{RGB}{50,135,60}
\providecommand{\redt}[1]{{\sethlcolor{hlred}\hl{#1}}}
\providecommand{\greent}[1]{{\sethlcolor{hlgreen}\hl{#1}}}
\providecommand{\rowbad}{\textcolor{markred}{$\times$}}
\providecommand{\rowgood}{\textcolor{markgreen}{$\checkmark$}}
\providecommand{\lightrule}{\noalign{\vskip 1.6pt{\color{black!22}\hrule height 0.3pt}\vskip 1.6pt}}
\providecommand{\dotrule}{\noalign{\vskip 1.6pt{\color{black!38}\hbox to \linewidth{\leaders\hbox to 2.6pt{\hss\rule[0pt]{0.6pt}{0.6pt}\hss}\hfill}}\vskip 1.6pt}}
\providecommand{\orig}[1]{{\sethlcolor{hlgray}\hl{#1}}}

\begin{tabularx}{\linewidth}{@{}l@{\hspace{3pt}}c>{\arraybackslash}X@{}}
\toprule
\multicolumn{3}{@{}>{\arraybackslash}p{\linewidth}@{}}{\textcolor{argumentcolor!75!black}{\textbf{Citation isolation}}\quad Shaded sentence relates CALM to nothing. \textcolor{markred}{$\times$ Slop-aware leaves CALM alone.} \textcolor{markgreen}{$\checkmark$ \methodName adds InfMem.}} \\
\lightrule
Original & & For language models, \orig{CALM [1] uses confidence-based early exiting at the token level}, achieving significant speedups on text generation. \\
\lightrule
Slop-aware & \rowbad & For language models, \redt{confidence-based methods like CALM [1] extend these layer-wise approaches to the token level}, achieving significant speedups during text generation\ldots \\
\dotrule
\methodName & \rowgood & For language models, CALM [1] uses confidence-based early exiting at the token level, achieving significant speedups on text generation, \greent{whereas InfMem [2] learns stopping policies at the chunk level for memory agents, achieving 3.3--5.1$\times$ speedup}. \\
\addlinespace[3pt]
\midrule
\multicolumn{3}{@{}>{\arraybackslash}p{\linewidth}@{}}{\textcolor{artifactcolor!85!black}{\textbf{Evidence gap}}\quad No concrete instance appears in the paper. \textcolor{markred}{$\times$ Slop-aware invents one.} \textcolor{markgreen}{$\checkmark$ \methodName states the gap.}} \\
\lightrule
Original & & \orig{\textit{No example input, output, case, or failure is displayed; the paper reports aggregate accuracies and agreement rates only.}} \\
\lightrule
Slop-aware & \rowbad & \redt{Worked Example. Consider a physical reasoning question: ``A rubber ball is dropped from rest and falls 5 meters\ldots'' Response A (preferred): ``The ball accelerates downward at approximately 9.8 m/s$^2$ due to gravity\ldots''}\ldots\ \textit{Nothing in it occurs in the manuscript or the project records.} \\
\dotrule
\methodName & \rowgood & \greent{This manuscript reports aggregate results from experiments but does not display concrete individual instances such as example inputs, outputs, or specific failure cases. The claims regarding model behavior and consistency gating effects rest on aggregate scores only.} \textit{Review reverts the editor's own inserted example} \redt{(e.g., consistent selection of option A)} \textit{as beyond the record.} \\
\addlinespace[3pt]
\midrule
\multicolumn{3}{@{}>{\arraybackslash}p{\linewidth}@{}}{\textcolor{argumentcolor!75!black}{\textbf{Argument graph}}\quad Shaded claim precedes its method. \textcolor{markred}{$\times$ Slop-aware keeps the order.} \textcolor{markgreen}{$\checkmark$ \methodName moves it after the method.}} \\
\lightrule
Original & & \orig{This runtime feedback can guide diffusion models to target repair at semantically relevant regions}\ldots\ $\downarrow$\ We introduce an execution-trace guided remasking pipeline that uses runtime diagnostics\ldots\ to localize failures and target repair in diffusion code generation. \\
\lightrule
Slop-aware & \rowbad & \redt{This runtime feedback can guide diffusion models to target repair at semantically relevant regions}\ldots\ $\downarrow$\ We introduce an execution-trace guided remasking pipeline \redt{described in Section [Method]}\ldots\ to localize failures and target repair in diffusion code generation. \\
\dotrule
\methodName & \rowgood & We introduce an execution-trace guided remasking pipeline that uses runtime diagnostics\ldots\ to localize failures and target repair in diffusion code generation. \greent{This runtime feedback can guide diffusion models to target repair at semantically relevant regions}\ldots \\
\addlinespace[3pt]
\midrule
\multicolumn{3}{@{}>{\arraybackslash}p{\linewidth}@{}}{\textcolor{artifactcolor!85!black}{\textbf{Figure exposition}}\quad Shaded diagram text restates the paper. \textcolor{markred}{$\times$ Slop-aware leaves the image.} \textcolor{markgreen}{$\checkmark$ \methodName erases six of eight phrases.}} \\
\lightrule
Original & & \orig{Stage 1: Rollout Generation} \textperiodcentered\ \orig{Stage 2: Metamorphic Audit} \textperiodcentered\ \orig{Stage 3: Trust Scoring} \textperiodcentered\ \orig{[agent icon] = Agent; $\times$ = Outcome} \textperiodcentered\ \orig{Legend:} \textperiodcentered\ \orig{Orange box = State-preserving command} \textperiodcentered\ \orig{Purple box = Distance computation \& Scoring} \textperiodcentered\ \orig{Green box = World Model} \\
\lightrule
Slop-aware & \rowbad & \textit{The baseline edits source text only, so every phrase above remains in the image.} \\
\dotrule
\methodName & \rowgood & \greent{Stage 1: Rollout Generation} \textperiodcentered\ \greent{Stage 2: Metamorphic Audit} \textperiodcentered\ \greent{Stage 3: Trust Scoring} \textperiodcentered\ \greent{[agent icon] = Agent; $\times$ = Outcome} \textperiodcentered\ \redt{Legend:} \textperiodcentered\ \greent{Orange box = State-preserving command} \textperiodcentered\ \greent{Purple box = Distance computation \& Scoring} \textperiodcentered\ \redt{Green box = World Model}\quad \textit{Green phrases are erased in round 1 and review confirms that nothing outside their rectangles changed; red phrases are not located and remain.} \\
\bottomrule
\end{tabularx}

\caption{One passage per remaining pattern before revision, after slop-aware revision, and after \MitigateName, marking \orig{the located slop}, \redt{what a revision lost or left}, and \greent{what it kept or repaired}. Brackets replace citation and reference commands, and the arrow marks the order of two statements across intervening text. Manuscript scores (original / slop-aware / \MitigateName): \mbox{0.615 / 0.385 / 0.571}, \mbox{1.000 / 0.000 / 0.500}, \mbox{0.333 / 0.167 / 0.000}, and \mbox{0.500 / 0.500 / 0.167}, where the baseline edits no images and so keeps the original figure score.}
\label{tab:revision-cases-more}
\end{table}

\end{document}